\documentclass[letterpaper, 10 pt, conference]{ieeeconf}  

\IEEEoverridecommandlockouts                              

\usepackage{graphicx}
\usepackage{graphics} 
\usepackage{amsmath} 
\usepackage{amssymb}  
\usepackage{makecell}
\usepackage{booktabs}
\usepackage{multirow}
\usepackage{xcolor}
\usepackage{cancel}
\usepackage{hyperref}
\hypersetup{
    colorlinks=true,
    linkcolor=blue,
    filecolor=magenta,      
    urlcolor=cyan,
}

\usepackage{caption}
\usepackage{subcaption}
\usepackage[ruled,linesnumbered]{algorithm2e} 
\usepackage{multicol}

\usepackage[absolute]{textpos}
\title{\LARGE \bf
Memory-based Deep Reinforcement Learning for POMDPs
}

\author{Lingheng Meng$^{1}$, Rob Gorbet$^{2}$ and Dana Kuli\'c$^{3}$
\thanks{$^{1}$Lingheng Meng is with Department of Electrical and Computer Engineering, University of Waterloo, 200 University Avenue West, Waterloo, ON, Canada
        {\tt\small lingheng.meng@uwaterloo.ca}}%
\thanks{$^{2}$Rob Gorbet is with Departments of Knowledge Integration and Electrical and Computer Engineering, University of Waterloo, 200 University Avenue West, Waterloo, ON, Canada
        {\tt\small rob.gorbet@uwaterloo.ca}}%
\thanks{$^{3}$Dana Kuli\'c is with the Faculty of Engineering, Monash University, 14 Alliance Lane, Melbourne, Victoria, Australia
        {\tt\small dana.kulic@monash.edu}}%
}

\begin{document}

\begin{textblock}{5}(2.5,0.6)
\parbox{16cm}{\color{gray}\Large \centering This paper has been accepted for publication at the 2021 IEEE/RSJ International Conference on Intelligent Robots and Systems (IROS 2021) in Prague, Czech Republic. Copyright may be transferred.}
\end{textblock}

\maketitle
\thispagestyle{empty}
\pagestyle{empty}

\begin{abstract}
A promising characteristic of Deep Reinforcement Learning (DRL) is its capability to learn optimal policy in an end-to-end manner without relying on feature engineering. However, most approaches assume a fully observable state space, i.e. fully observable Markov Decision Processes (MDPs). In real-world robotics, this assumption is unpractical, because of issues such as sensor sensitivity limitations and sensor noise, and the lack of knowledge about whether the observation design is complete or not. These scenarios lead to Partially Observable MDPs (POMDPs). In this paper, we propose Long-Short-Term-Memory-based Twin Delayed Deep Deterministic Policy Gradient (LSTM-TD3) by introducing a memory component to TD3, and compare its performance with other DRL algorithms in both MDPs and POMDPs. Our results demonstrate the significant advantages of the memory component in addressing POMDPs, including the ability to handle missing and noisy observation data.
\end{abstract}

\section{Introduction}
Deep Reinforcement Learning (DRL) \cite{mnih2013playing,lillicrap2015continuous}has been intensively studied in simulated environments, such as games \cite{mnih2013playing} and simulated robots \cite{lillicrap2015continuous}, as well as in real-world studies, such as robotics control \cite{gu2017deep,zhang2015towards,tai2017virtual} and human-robot interaction \cite{qureshi2016robot, chen2017socially, 10.1145/3408876}. DRL enables end-to-end policy learning on tasks with high-dimensional state and action spaces, without relying on  labour-consuming feature engineering. However, most works focus on developing algorithms \cite{mnih2013playing,lample2017playing,lillicrap2015continuous,schulman2015trust,schulman2017proximal,fujimoto2018addressing,haarnoja2018soft} for Markov Decision Processes (MDPs) with fully observable state spaces \cite{duan2016benchmarking}, i.e. the observation at each time step fully represents the state of the environment.  Few works consider the more complex Partially Observable Markov Decision Process (POMDP) where the observation is just a partial representation of the underlying state. However, POMDPs are ubiquitous in real robotics applications \cite{murphy2000survey}, such as robot navigation \cite{candido2011minimum}, robotic manipulation \cite{pajarinen2017robotic}, autonomous driving \cite{talpaert2019exploring,wang2019autonomous}, and planning under uncertainty \cite{wang2019q,cheng2018reinforcement}. Partial observability may be due to limited sensing capability, or an incomplete system model resulting in uncertainty about full observability. 

\begin{figure}
    \centering
    \includegraphics[width=0.8\linewidth]{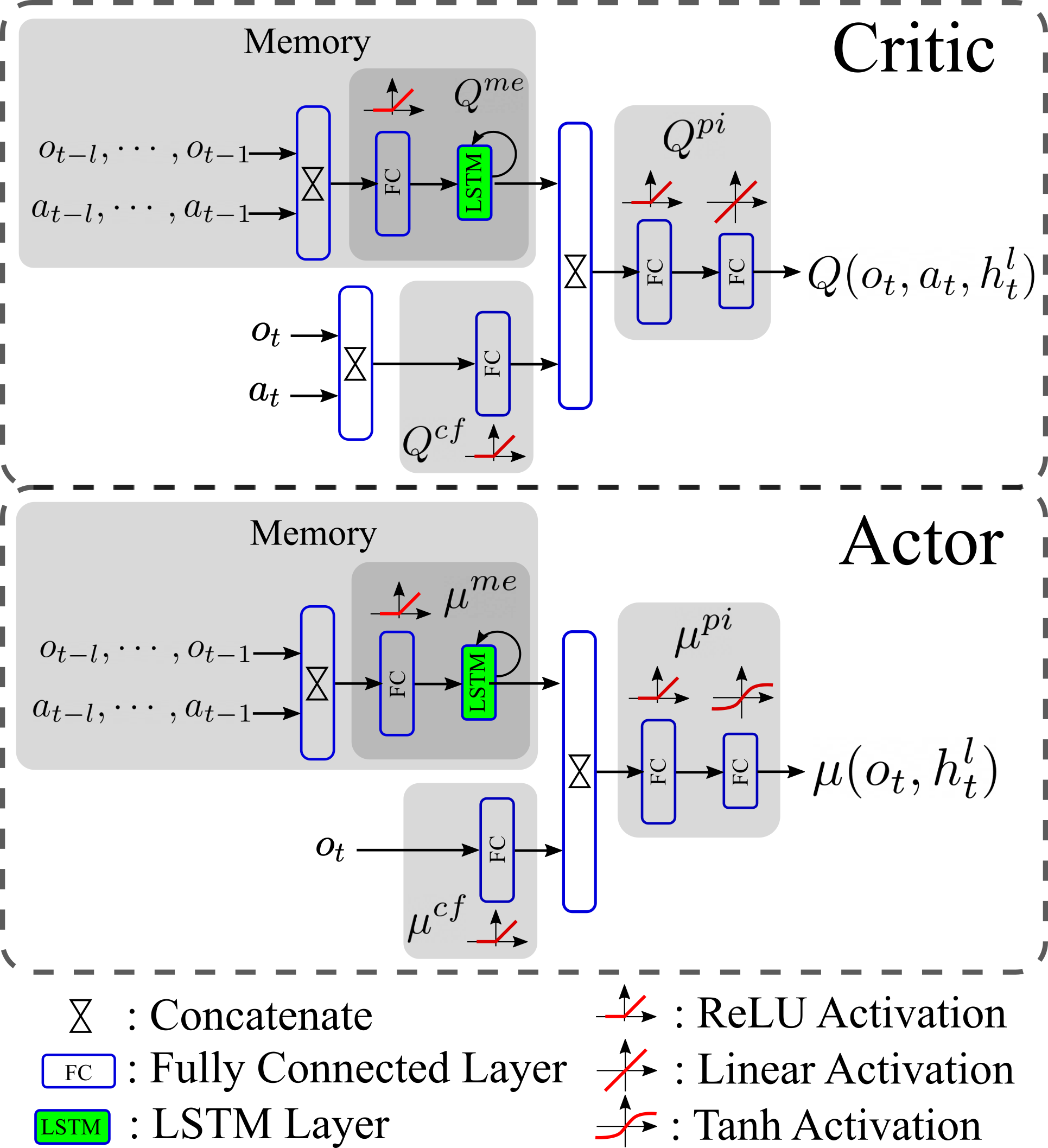}
    \caption{Recurrent Actor-Critic Framework}
    \label{fig:Recurrent_Actor-Critic_Framework}
\end{figure}

POMDPs have been tackled with the concept of belief state \cite{russell2002artificial}, which represents the agent's current belief about the possible physical states it might be in, given the sequence of actions and observations up to that point. These algorithms are designed to estimate the belief state, then the value function and/or the policy are learned based on the belief state \cite{shani2013survey}. However, these methods need to know the environment model and the state space and they only work on tasks with small state and action spaces. 

POMDPs have also been addressed with DRL, for both discrete \cite{hausknecht2015deep,lample2017playing,zhu2017improving} and continuous \cite{song2018recurrent,wang2019autonomous} control problems. Recurrent Neural Networks (RNN) have been exploited in DRL to solve POMDPs by considering both the current observation and action, and the history of the past observations and actions \cite{heess2015memory,song2018recurrent,zhang2016learning,zhu2017improving,lample2017playing}. 

In this paper, we propose a memory-based DRL, called \textit{Long-Short-Term-Memory-based Twin Delayed Deep Deterministic Policy Gradient (LSTM-TD3)}, for continuous robot control.  We provide a comparison study with other DRL algorithms where both MDP and POMDP versions of the tasks are investigated to demonstrate how observeability properties influence performance on both POMDPs and MDPs. Compared to other DRL algorithms, results show that LSTM-TD3 improves performance significantly on POMDPs where the observation space is disturbed to reduce the observability. We will also provide an ablation study to show the contribution of each design component, and discuss the advantages and disadvantages of the proposed method.

\section{Related Work}
\label{sec:headings}

A number of previous works have investigated how memory can be incorporated into DRL.  Deep Recurrent Q-Learning (DRQN) \cite{hausknecht2015deep} adds recurrency to the Deep Q-Network (DQN) \cite{mnih2015human} by replacing the first post-convolutional fully-connected layer with a recurrent Long-Short-Term-Memory (LSTM). The results on  Atari 2600 games show DRQN significantly outperforms DQN on POMDPs, which validates the effectiveness of memory extracted by the LSTM for solving POMDPs. \cite{lample2017playing} investigated a similar idea but augmented the structure with an auxiliary game feature, e.g. presence of enemies, learning in 3D environments in first-person shooter games. The results show the proposed architecture substantially outperforms DRQN. These methods only consider past observations in the history.  \cite{zhu2017improving} proposed Action-specific Deep Recurrent Q-Network (ADRQN) to also consider past actions in the memory. However, these works are based on tasks with discrete action spaces, rather than on continuous control tasks.

\cite{heess2015memory} extended Deterministic Policy Gradient (DPG) \cite{silver2014deterministic} to Recurrent DPG (RDPG) by adding LSTM and investigated it on continuous control tasks with partial observations. Dramatic performance improvement was observed with memory. However, even though the observation space was large for some tasks, the action space had relatively few dimensions for the investigated tasks. \cite{song2018recurrent} investigated RDPG on bipedal locomotion tasks with both visual and sensory input, but only one task was examined. Different from directly optimizing RNN, \cite{zhang2016learning} proposed to augment the observations and actions with the continuous memory states and use guided policy search to optimize a linear policy. The method shows better performance than other policy search methods. However, the guided policy search is less powerful and generalizable than non-linear policy.

In our work, we consider continuous control tasks with large observation and action spaces and propose LSTM-TD3 within a recurrent actor-critic framework, which is a further improvement of RDPG by exploiting  TD3 to reduce the overestimation problem.

\section{Background}

\subsection{Decision Process}
A Markov Decision Process (MDP) is a sequential decision process for a fully observable, stochastic environment with a Markovian transition model and additive rewards \cite{bellman1957markovian,russell2002artificial}. Formally, MDP can be defined as a 4-tuple $(S,A,P,R)$, where $S$ is the state space, $A$ is the action space, $P$ is the transition probability and $R$ is the reward function. At each discrete time $t$, an agent selects an action $a_t\in A$ in state $s_t \in S$, transitions to the next state $s_{t+1}$ with probability $P(s_{t+1}\mid s_t, a_t)$, and receives the immediate reward $R(s_t, a_t, s_{t+1})$. Partially Observable Markov Decision Process (POMDP) \cite{aastrom1965optimal,burnetas1997optimal,russell2002artificial} is a generalization of a MDP, but does not assume that the state is fully observable, and is defined as a 6-tuple $(S,A,P,R,O,\Omega)$, where $S$, $A$, $P$, and $R$ are the same as that in MDP, with an additional observation space $O$ and observation model $\Omega$. Although the underlying state transition in a POMDP is the same as those in an MDP, the agent cannot observe the underlying state, instead it receives an observation $o_{t+1} \in O$ when reaching the next state $s_{t+1}$ with the probability $\Omega(o_{t+1} \mid s_{t+1})$.

The goal for an agent in either MDPs or POMDPs is to choose actions at each time step that maximize its expected future discounted return $\mathbb{E}\left [ \sum_{t=0}^{\infty } \gamma^t r_t \right ]$, where $r_t$ is the immediate reward received at time $t$ and $\gamma \in \left [ 0,1 \right ]$ is the discount factor that describes the preference of the agent for current rewards over future rewards. 

\subsection{Reinforcement Learning (RL)}
RL \cite{sutton2018reinforcement} solves decision problems such as MDPs and POMDPs based on a learning paradigm where an agent learns to act by trial-and-error without knowing the underlying transition and reward model. Specifically, at a discrete time $t$ an agent interacts with the external environment by taking action $a_t$ according to either a stochastic policy $\pi (a_t \mid s_t)$ or a deterministic policy $a_t=\mu(s_t)$ when observing the current state $s_t$. By continuously interacting with the environment and receiving new states and rewards, an agent learns an optimal policy to maximize the expected future return. Accompanying the policy $\pi$ or $\mu$, it is common to learn a state-value function $V(s)$ and/or an action-value function $Q(s,a)$. For MDP, The state-value function $V^{\pi}(s)$ of a state $s$ under a policy $\pi$ represents the expected return starting from $s$ and following $\pi$ thereafter, and the action-value function $Q^{\pi}(s,a)$ of taking action $a$ in state $s$ under a policy $\pi$ is the expected return starting from $s$, taking $a$, and following $\pi$ thereafter. For POMDP, since the state $s$ is not observable, the observation $o$ is used in learning these value functions or policy. Traditional RL methods employ either tabular representation or simple function approximation, with hand-crafted feature construction, to represent the value function, while DRL \cite{mnih2013playing} \cite{lillicrap2015continuous} learns the value function and policy in an end-to-end manner without feature engineering by recruiting deep neural networks.

\subsection{Long-Short-Term-Memory (LSTM)}
LSTM \cite{hochreiter1997long} is a type of Recurrent Neural Network (RNN) \cite{goodfellow2016deep} that has an outer recurrence from the outputs to the inputs of the hidden layer and also an internal recurrence between LSTM-Cells. Within a LSTM-Cell, a system of gating units control the flow of information, and enable the remembering and forgetting of information given a sequence of inputs. 

\subsection{Twin Delayed Deep Deterministic Policy Gradient (TD3)}
TD3 \cite{fujimoto2018addressing} is a variant of Deep Deterministic Policy Gradient (DDPG) \cite{lillicrap2015continuous} designed to address the overestimation problem \cite{thrun1993issues} in Actor-Critic methods. Specifically, TD3 employs two critics $Q_{1}$ and $Q_{2}$, and uses the minimum of the predicted optimal future return in state $s_{t+1}$ to bootstrap the Q-value of the current state $s_t$ and action $a_t$. 

\section{Proposed Approach}
In this paper, we propose a memory-based DRL algorithm named LSTM-TD3 within a recurrent actor-critic framework, where both the actor and the critic employ recurrent neural networks, as illustrated in Fig. \ref{fig:Recurrent_Actor-Critic_Framework}. In this section, we will first introduce the proposed recurrent actor-critic framework, then present the optimization method for the actor-critic.

In the proposed approach,  a mini-batch of $N$ experiences $\left \{ (h_{t}^{l}, o_t, a_t,  r_t, o_{t+1}, d_t)_{i} \right \}_{i=1}^{N}$ is sampled from the replay buffer $D$ of experiences $(o_t, a_t, r_t, o_{t+1}, d_t)$, where $d_t$ indicates whether the terminal state is reached after observing $o_{t+1}$ and for each sample the past history $h_{t}^{l}$ with length $l$ until observation $o_t$ at time $t$ is defined as:
\begin{equation}
    h_{t}^{l} = \left\{\begin{matrix}
                o_{t-l}, a_{t-l}, \cdots , o_{t-1}, a_{t-1} & \text{if } l,t \geq 1.\\ 
                o^{0}, a^{0} & \text{Otherwise}.
                \end{matrix}\right.
    \label{eq:history_definition}
\end{equation}
where $o^{0}$ and $a^{0}$ are the zero-valued dummy observation and action vectors with the same dimensions as those of the normal observation and action. As defined in Eq. \ref{eq:history_definition}, if history length $l \geq 1$ and time step $t\geq 1$, the history $h_{t}^{l}$ at time $t$ is defined as the past $l$ (observation, action) pairs, otherwise no history is used and zero-valued dummy vectors (observation, action) are used as input to the memory component. 

\subsection{Recurrent Actor-Critic Framework}
\label{subsec:Recurrent_Actor_Critic_Framework}
The structure of the proposed recurrent actor-critic framework is illustrated in Fig. \ref{fig:Recurrent_Actor-Critic_Framework}, where Long-Short-Term-Memory is introduced to extract information beneficial to the actor and critic from past history. The proposed framework can handle history of any length.

Formally, given a mini-batch sample of experiences, the memory-based critic Q, as illustrated in Fig. \ref{fig:Recurrent_Actor-Critic_Framework}, can be seen as a compound function of the memory extraction $Q^{me}$, the current feature extraction $Q^{cf}$, and the perception integration $Q^{pi}$ components, following Eq. \ref{eq:memory_based_critic}
\begin{equation}
    \begin{aligned}
        Q(o_t, a_t, h_t^l) &= Q^{me} \circ Q^{cf} \circ Q^{pi}\\
                           &= Q^{pi}(Q^{me}(h_t^l) \Join Q^{cf}(o_t,a_t))
    \end{aligned}
    \label{eq:memory_based_critic}
\end{equation}
where $\Join$ indicates the concatenation operation, $Q^{me}$ is the extracted memory based on history $h_t^l$, and $Q^{cf}$ is the extracted current feature based on current observation $o_t$ and action $a_t$. 

Similarly, the memory-based actor $\mu$ is also a compound function of the memory extraction $\mu^{me}$, the current feature extraction $\mu^{cf}$, and the perception integration $\mu^{pi}$ components, defined as follows:
\begin{equation}
    \begin{aligned}
        \mu(o_t, h_{t}^{l}) &= \mu^{me} \circ \mu^{cf} \circ \mu^{pi} \\
                            &=\mu^{pi}(\mu^{me}(h_t^l)\Join \mu^{cf}(o_t))
    \end{aligned}
    \label{eq:memory_based_actor}
\end{equation}
where $\mu^{me}$ is the extracted memory based on history $h_t^l$, and $\mu^{cf}$ is the extracted current feature based on current observation $o_t$.

\subsection{Optimization of the Recurrent Actor-Critic}
The optimization of the proposed recurrent actor-critic framework follows that of TD3. Specifically, each critic $Q_{j\in \begin{Bmatrix} 1,2 \end{Bmatrix}}$ is optimized to minimize the mean-square-error between the predicted $Q_j$ and the estimated target $\hat{Q}$ with respect to the parameters $\theta^{Q_{j}}$ of the critic $Q_j$, as follows:
\begin{equation}
    \text{min}_{\theta^{Q_j}} \text{ } \mathbb{E}_{\left \{ (h_{t}^{l}, o_t, a_t, r_t, o_{t+1}, d_t)_{i} \right \}_{i=1}^{N}}(Q_j - \hat{Q})^2
    \label{eq:optimize_critic}
\end{equation}
where given the definition of memory-based critic (Eq. \ref{eq:memory_based_critic}) and actor (Eq. \ref{eq:memory_based_actor}), the target Q-value $\hat{Q}$ based on the target actor $\mu^{-}$ and critic $Q_j^{-}$ is defined as Eq. \ref{eq:target_Q}
\begin{equation}
    \hat{Q} = r_t + \gamma \ast (1-d_t) \ast \underset{j=1,2}{\min} Q_{j}^{-}(o_{t+1}, a^-, h_{t+1}^{l})
    \label{eq:target_Q}
\end{equation}
where $a^-=\mu^{-}(o_{t+1}, h_{t+1}^{l})+\epsilon$ with $\epsilon \sim \text{clip}(\mathbb{N}(0, \sigma), -c, c)$ and $c$ is the boundary of target action noise, $h_{t+1}^{l} = (h_{t}^{l}- (o_{t-l}, a_{t-l})) \cup (o_t, a_t) $ is the $l$ observation and action pairs before $o_{t+1}$, and the minimum of the estimated optimal Q-values of the two target critics in $(o_{t+1}, h_{t+1}^{l})$ is taken to bootstrap the target Q-value of $(o_t, a_t, h_{t}^{l})$.

For the actor, its parameters $\theta^\mu$ are optimized to maximize the approximated Q-value in observation $(o_{t}, h_{t}^{l})$ and the corresponding estimated optimal action $\mu(o_t, h_{t}^{l})$ with respect to the parameters of the actor, as follows:
\begin{equation}
    \text{max}_{\theta^{\mu}} \text{ } \mathbb{E}_{\left \{ (h_{t}^{l}, o_t)_{i} \right \}_{i=1}^{N}} Q(o_t, \mu(o_t, h_{t}^{l}), h_{t}^{l})
    \label{eq:optimize_actor}
\end{equation}
where the $Q$ could be either of the two critics $Q_1$ and $Q_2$, as in TD3. The pseudo-code for optimizing the recurrent actor-critic can be found in Alg. \ref{alg:LSTM_TD3}.

\begin{figure}
    \centering
    \includegraphics[width=.99\linewidth]{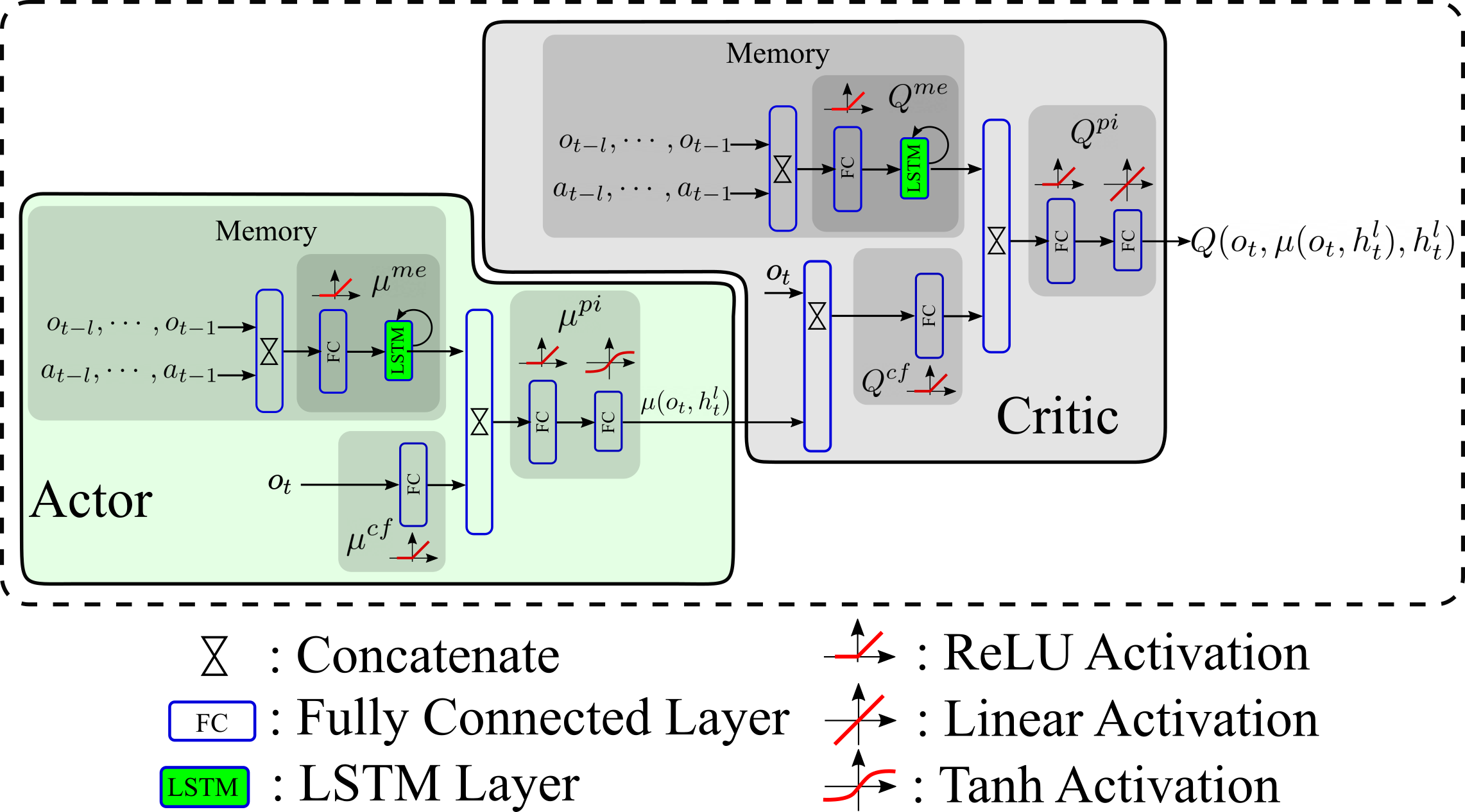}
    \caption{Actor Optimization, where the parameters $\theta^{Q_1}$ of the critic $Q_1$ is fixed while the parameters $\theta^{\mu}$ of the actor is optimized according to Eq. \ref{eq:optimize_actor}.}
    \label{fig:actor_optimization}
\end{figure}
\begin{algorithm}[ht!]
    \SetAlgoLined
    \SetNoFillComment
    \DontPrintSemicolon
    \KwIn{History length $L$}
    Initialize critics $Q_{\theta^{Q_1}}$, $Q_{\theta^{Q_2}}$, and actor $\mu_{\theta^{\mu}}$ with random parameters $\theta^{Q_1}$, $\theta^{Q_2}$ and $\theta^{\mu}$\;
    Initialize target networks $\theta^{Q_1^-} \leftarrow \theta^{Q_1}$, $\theta^{Q_2^-} \leftarrow \theta^{Q_2}$ and $\theta^{\mu^-} \leftarrow \theta^{\mu}$\;
    Initialize environment $o_1$ = env.reset(), past history $h_1^l \leftarrow \mathbf{0}$, and replay buffer $D$\;
    \For{$t=1$ \KwTo $T$}{
        \tcc{Interacting}
        Select action with exploration noise $a_t \sim \mu_{\theta^\mu}(o_t, h_t^l)+\epsilon$, $\epsilon \sim \mathbb{N}(0,\sigma)$\;
        Interact and observe new observation, reward, and done flag:
        $o_{t+1}$, $r_t$, $d_t$ = env.step($a_t$)\;
        Store experience tuple $(o_t, a_t, r_t, o_{t+1}, d_t)$ in $D$\;
        \eIf{$d$}{
            Reset environment $o_{t+1}$ = env.reset() and history $h_{t+1}^l \leftarrow \mathbf{0}$\;
        }{
            \tcc{Update $h_{t+1}^l$}
            $h_{t+1}^{l} = (h_t^{l}- (o_{t-l}, a_{t-l})) \cup (o_t, a_t) $ \;
        }
        \tcc{Learning}
        Sample mini-batch of $N$ experiences with their corresponding histories $\left \{ (h_{t}^{l}, o_t, a_t,  r_t, o_{t+1}, d_t)_{i} \right \}_{i=1}^{N}$ from $D$\;
        Optimize $Q_j$ according to Eq. \ref{eq:optimize_critic}\;
        Optimize $\mu$ according to Eq. \ref{eq:optimize_actor}\;
        Update target actor-critic
    }
    \caption{Pseudo-code for LSTM-TD3}
    \label{alg:LSTM_TD3}
\end{algorithm}

\section{Experiment Settings}
\label{sec:experiment}

\begin{figure}
    \begin{subfigure}[b]{.19\linewidth}
        \centering
        \includegraphics[width=1.5cm, height=1.5cm]{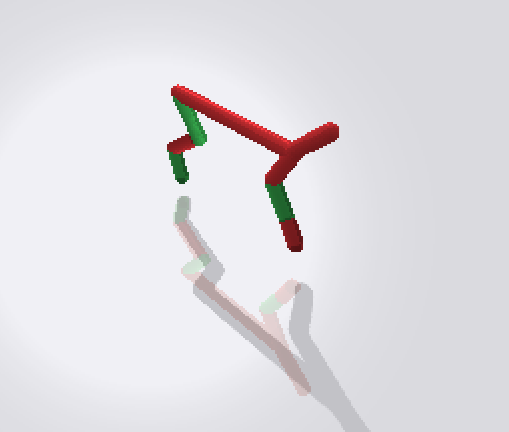}
        \caption{}
    \end{subfigure}
    \begin{subfigure}[b]{.19\linewidth}
        \centering
        \includegraphics[width=1.5cm, height=1.5cm]{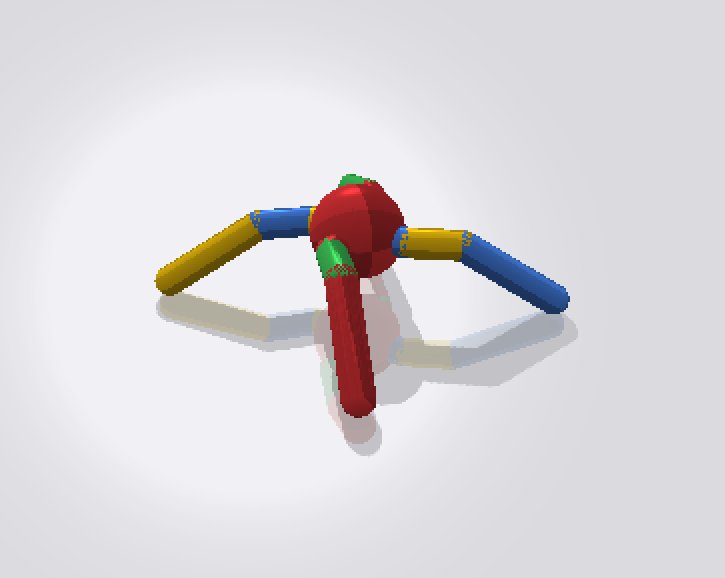}
        \caption{}
    \end{subfigure}
    \begin{subfigure}[b]{.19\linewidth}
        \centering
        \includegraphics[width=1.5cm, height=1.5cm]{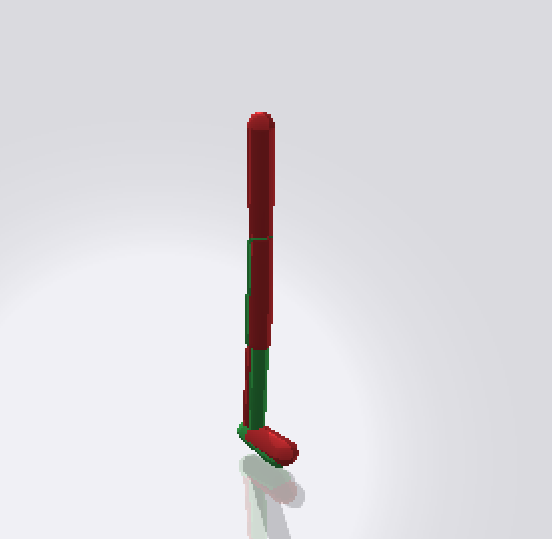}
        \caption{}
    \end{subfigure}
    \begin{subfigure}[b]{.19\linewidth}
        \centering
        \includegraphics[width=1.5cm, height=1.5cm]{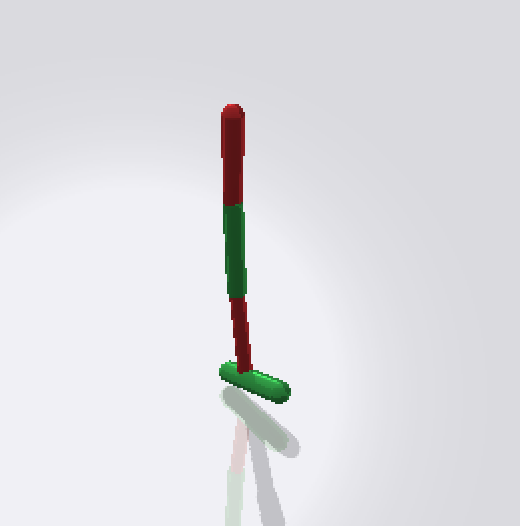}
        \caption{}
    \end{subfigure}
    \begin{subfigure}[b]{.19\linewidth}
        \centering
        \includegraphics[width=1.5cm, height=1.5cm]{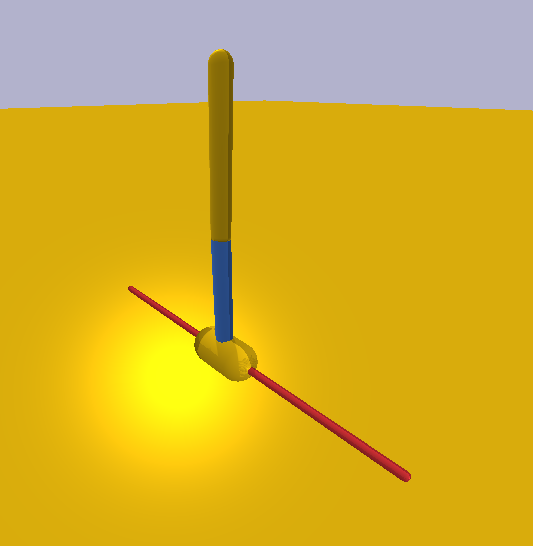}
        \caption{}
    \end{subfigure}%
    \caption{Example PyBulletGym tasks (a) HalfCheetahPyBulletEnv-v0, (b) AntPyBulletEnv-v0, (c) Walker2DPyBulletEnv-v0, (d) HopperPyBulletEnv-v0, and (e) InvertedDoublePendulumPyBulletEnv-v0.}
    \label{fig:Example_PyBulletGym_tasks}
\end{figure}

The tasks (Fig. \ref{fig:Example_PyBulletGym_tasks}) tested in this work come from PyBulletGym\footnote{https://github.com/benelot/pybullet-gym}, an open-source implementation of the OpenAI Gym MuJoCo environment based on BulletPhysics\footnote{https://github.com/bulletphysics/bullet3}. In this work, an MDP-version and 4 POMDP-versions of each task are investigated, described in Table \ref{tab:MDP_and_POMDP_version_of_Tasks}. The MDP-version is the original task, as it has a fully observeable state-space, while the 4 POMDP-versions simulate different scenarios that potentially cause partial observability in real applications. Specifically, the POMDP-RemoveVelocity (POMDP-RV) is designed to simulate the scenario where the observation space is not well-designed, which is applicable to a novel control task that is not well-understood by researchers and therefore the designed observation may not fully capture the underlying state of the robot. The POMDP-Flickering (POMDP-FLK) models the case where remote sensor data is lost during long-distance data transmission. The case when a subset of the sensors are lost is simulated in POMDP-RandomSensorMissing (POMDP-RSM). Sensor noise is simulated in POMDP-RandomNoise (POMDP-RN).

\begin{table}[ht]
 \caption{MDP- and POMDP-version of Tasks}
  \centering
  \begin{tabular}{lll}
    \toprule
    \textbf{Name}  & \textbf{Description}     & \makecell[l]{\textbf{Hyper}-\\\textbf{parameter}} \\
    \midrule
    MDP            & Original task & $-$     \\\hline
    POMDP-RV       & \makecell[l]{Remove all velocity-related entries in \\the observation space.} & $-$      \\\hline
    POMDP-FLK      & \makecell[l]{Reset the whole observation to 0 with \\probability $p_{flk}$.} & $p_{flk}$  \\\hline
    POMDP-RN       & \makecell[l]{Add random noise $\epsilon \sim \mathbb{N}(0, \sigma_{rn})$ \\to each entry of the observation.}   &  $\sigma_{rn}$            \\\hline
    POMDP-RSM      & \makecell[l]{Reset an entry of the observation to 0 \\with  probability $p_{rsm}$.}     & $p_{rsm}$              \\
    \bottomrule
  \end{tabular}
  \label{tab:MDP_and_POMDP_version_of_Tasks}
\end{table}


The baselines used to compare with the proposed LSTM-TD3 are the DDPG \cite{lillicrap2015continuous}, SAC \cite{haarnoja2018soft}, TD3 \cite{fujimoto2018addressing}, TD3 with Observation-Window (TD3-OW) where the $o_t$ is simply concatenated with the observations within the history window $h_{t}^{l}$ to form an observation as input, and TD3 with Observation-Window-AddPastAct (TD3-OW-AddPastAct) where $o_t$ is concatenated with the observations and the actions within the history window $h_{t}^{l}$. The hyperparameters for the baseline algorithms were always the defaults provided in OpenAISpinningUp\footnote{https://spinningup.openai.com}. For the proposed algorithm, hyperparameters were empirically set to that for TD3, and the network structures of the LSTM-TD3 were chosen to have a similar number of parameters to the networks in TD3. All reported results are averaged over 10 evaluation episodes based on 4 different random seeds. The code used for this work can be found in \url{https://github.com/LinghengMeng/LSTM-TD3}. All hyperparameter testing and additional results (e.g. LSTM-TD3 in POMDPs with lower observability and larger history length than that reported here) are reported in the Supplementary Material (available at \url{http://arxiv.org/abs/2102.12344}).

\section{Results}

\subsection{Performance Comparison}
\label{subsec:performance_comparison}
\begin{figure*}[t]
    \centering
    \includegraphics[width=.95\linewidth]{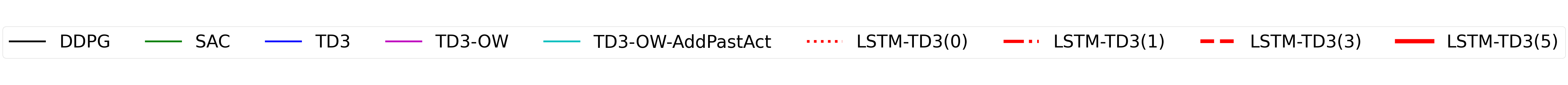}
    \subfloat{\includegraphics[width=.99\linewidth]{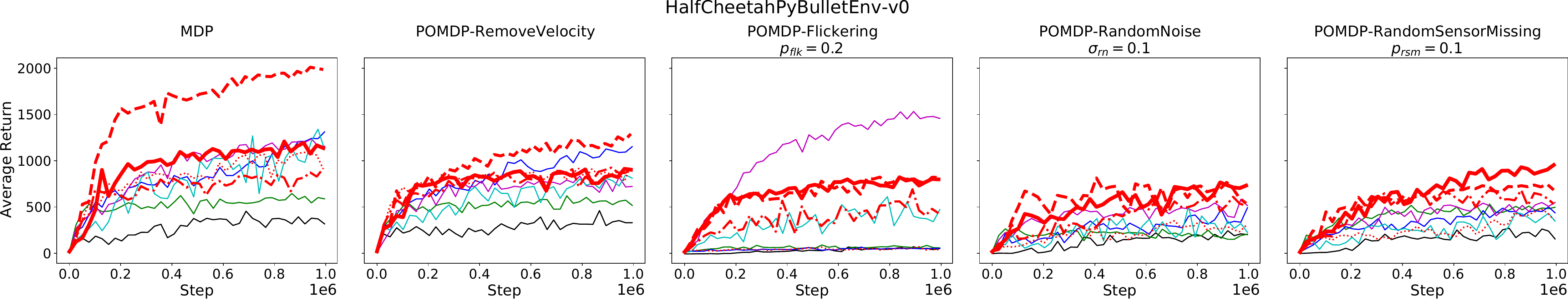}}\\\vspace{5px}
    \subfloat{\includegraphics[width=.99\linewidth]{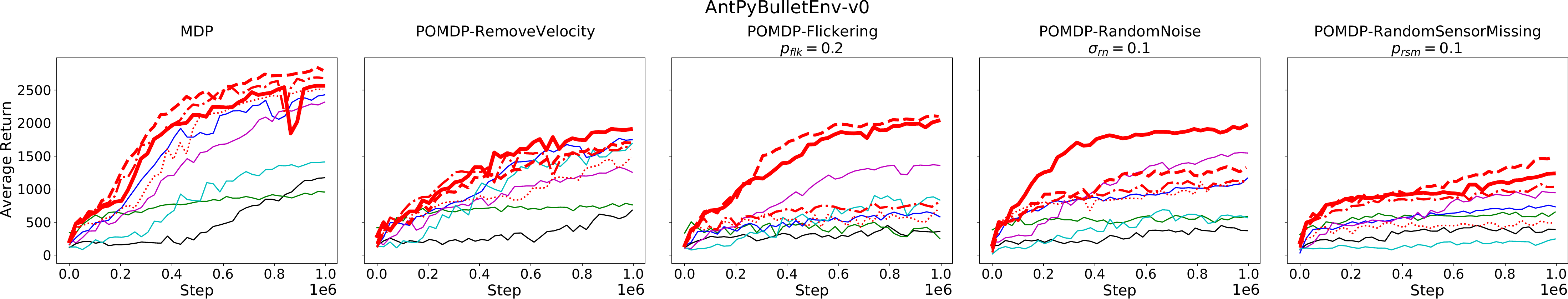}}\\\vspace{5px}
    \subfloat{\includegraphics[width=.99\linewidth]{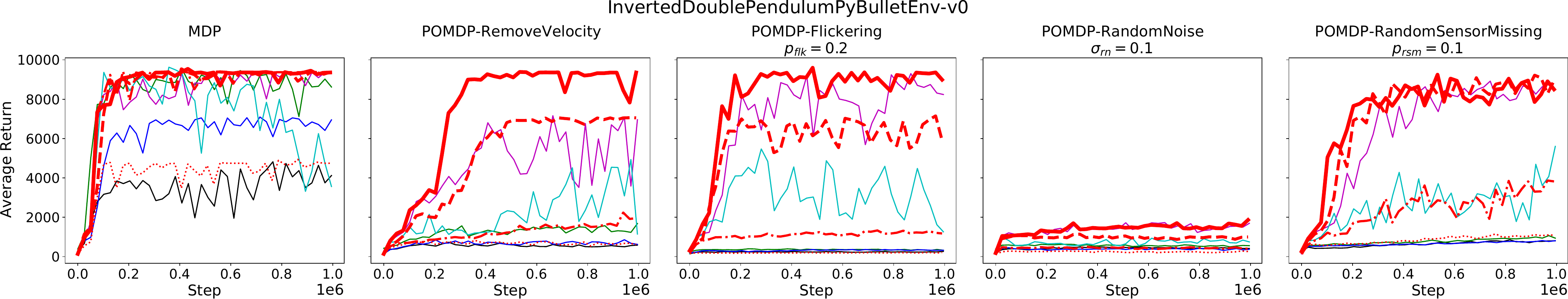}}
    \caption{Learning curves for PyBulletGym tasks, where to ease the comparison only average values are plotted. In the legend, the value in the bracket of LSTM-TD3 indicate the length of the history, e.g. LSTM-TD3(5) uses history length 5.}
    \label{fig:learning_curves_for_performance_comparison}
\end{figure*}

The rows of Fig. \ref{fig:learning_curves_for_performance_comparison} show the learning curves of the three sampled tasks, where the fist column shows the performance on MDP, while the following 4 columns show results on POMDPs. The results on MDP show that the proposed method has competitive performance to the baselines.  The results on the POMDPs highlight the advantage of having memory when solving partially observable tasks. On all types of POMDP, LSTM-TD3 outperforms all baselines, except on POMDP-RV of HalfCheetahPyBulletEnv-v0, where LSTM-TD3(5) shows slightly worse performance than TD3. Although TD3-OW shows better performance than DDPG, TD3, and SAC on POMDPs for most tasks, it still fails for some POMDPs, such as the POMDP-FLK version of most tasks. This reveals that simply concatenating observations is not a good choice, compared to having a LSTM-based memory extraction component as that in LSTM-TD3. LSTM-TD3(0) 
seems sensitive to random seeds (1st panel of the 3rd row of Fig. \ref{fig:learning_curves_for_performance_comparison}) as it achieves lower performance compared to that of TD3 and LSTM-TD3 with history length larger than 0. To explain this, even though we set the history for it to zero, it may still predict nonzero for the $Q^{me}$ (introduced in Eq. \ref{eq:memory_based_critic}), because the gradients with respect to the randomly initialized weights of  $Q^{me}$ may be nonzero and back-propagated, which could influence the agent during learning.

Particularly, a significant performance gap can be observed on POMDP-FLK for all tasks (the 3rd column in Fig. \ref{fig:learning_curves_for_performance_comparison}), where the baselines basically fail while LSTM-TD3 achieves comparative performance to that on MDP. This is especially promising for tasks where whole sensor data may be lost, either caused by hardware failure or by temporary occlusion, etc. Similar, dramatic performance improvement can be seen on POMDP-RN and POMDP-RSM.  

Surprisingly, comparing LSTM-TD3 and TD3, memory does not always help for POMDP-RV (the 2nd column in Fig. \ref{fig:learning_curves_for_performance_comparison}) of HalfCheetahPyBulletEnv-v0 and AntPyBulletEnv-v0. Intuitively, if the velocity is important to learn a task, there is no way to infer such information without past observations i.e. memory. However, if previous observations are available, the velocity can be inferred by differences in position between consecutive steps. This intuition can be clearly observed on the POMDP-RV of HopperPyBulletEnv-v0 and InvertedDoublePendulumPyBulletEnv-v0, where the performance of LSTM-TD3 is significantly better than that of TD3. For the HalfCheetahPyBulletEnv-v0 and AntPyBulletEnv-v0, if we compare the performance on MDP and POMDP-RV, we can still see a noticeable gap, which means velocity does contribute to learn a good policy. We hypothesize that LSTM-TD3(5) does not outperform TD3 on the POMDP-RV version of HalfCheetahPyBulletEnv-v0 and AntPyBulletEnv-v0 due to the fact that within the history window all speeds are very similar and velocity cannot be accurately inferred,  which may be caused by relatively high sampling rate.

\begin{figure*}
    \centering
    \includegraphics[width=.7\linewidth]{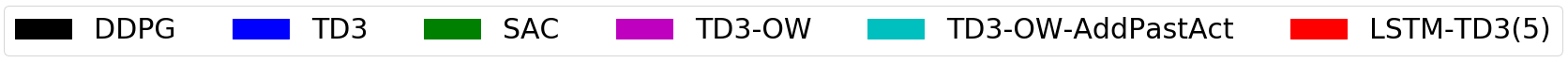}
    \includegraphics[width=.99\linewidth]{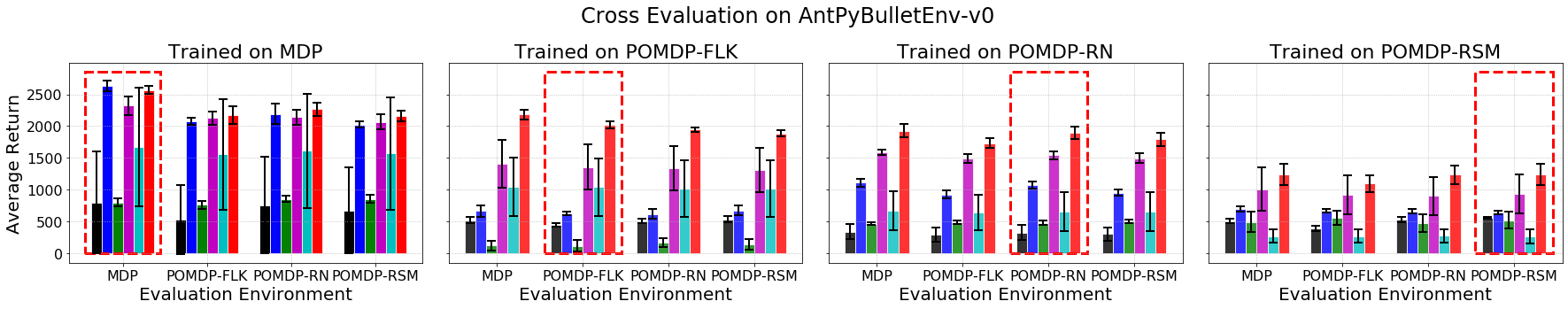}
    \caption{Cross Evaluation. In each panel, the x-axis indicates the evaluation environment (proposed in Table \ref{tab:MDP_and_POMDP_version_of_Tasks}) and the y-axis is the average return, where the bars highlighted with red dashed box correspond to performances evaluated on the same environment where the policies are trained, and errorbar on the bar tips indicates the standard deviation of the performance. }
    \label{fig:cross_evaluation}
\end{figure*}

Interestingly, by comparing the results of TD3-OW and TD3-OW-AddPastAct, we found that adding past actions consistently harms the performance compared to TD3-OW, which does not have past actions in its observation window. Even though TD3-OW-AddPastAct still outperforms TD3 on POMDP, it performs worse than TD3 on MDP, which is undesirable if we have no prior knowledge of whether the current design of the observation space is partially or fully observable. Ideally, even if the past action-related information does not provide anything new beyond the past observation, it can be safely ignored and should not harm the performance. We think this is related to the simple construction method where actions are concatenated with observations to form a single observation that includes history information. In this way, the observation dimension is expanded, which makes the learning harder. In addition, this simple construction method treats all observations equally instead of prioritizing the most recent observation, which is normally more valuable in decision-making than earlier observations. This observation based on TD3-OW and TD3-OW-AddPastAct in fact supports our idea to structurally separate the memory extraction and current feature extraction in the recurrent actor-critic framework (Fig. \ref{fig:Recurrent_Actor-Critic_Framework}) designed for LSTM-TD3, then combine them together to further learn a presentation of the critic and the actor. In section \ref{subsec:ablation_add_past_action}, we will further investigate if adding past actions is beneficial for LSTM-TD3.

\subsection{Policy Generalization}
To better understand the generalization of the learned policy using LSTM-TD3, we evaluated the learned policy on a different version of a given task, e.g. if the policy is learned on the MDP-version of a task, and evaluated on the POMDP-versions of the task. This is valuable for real applications where the environment may be non-stationary. Fig. \ref{fig:cross_evaluation} shows the cross evaluation results on AntPyBulletEnv-v0. POMDP-RV is not included as it has a different observation dimension which corresponds to a different input shape for the neural networks. From the first panel (i.e. policies trained on MDP), TD3, TD3-OW, and LSTM-TD3 significantly outperform DDPG, SAC, and TD3-OW-AddPastPact. When evaluating on POMDPs, there is always a decrease for TD3, TD3-OW, and LSTM-TD3, but LSTM-TD3 is the most robust and achieves better performance on these evaluation environments than TD3 and TD3-OW. As for the last three panels, even though LSTM-TD3 still outperforms TD3-OW significantly when evaluated on a different environment, for each algorithm there is not much change in performance. Actually, when trained on POMDP-FLK and evaluated on MDP, LSTM-TD3 achieves a better performance than evaluated on POMDP-FLK.

\begin{figure}
    \begin{subfigure}[t]{.49\linewidth}
        \centering
        \includegraphics[width=\linewidth]{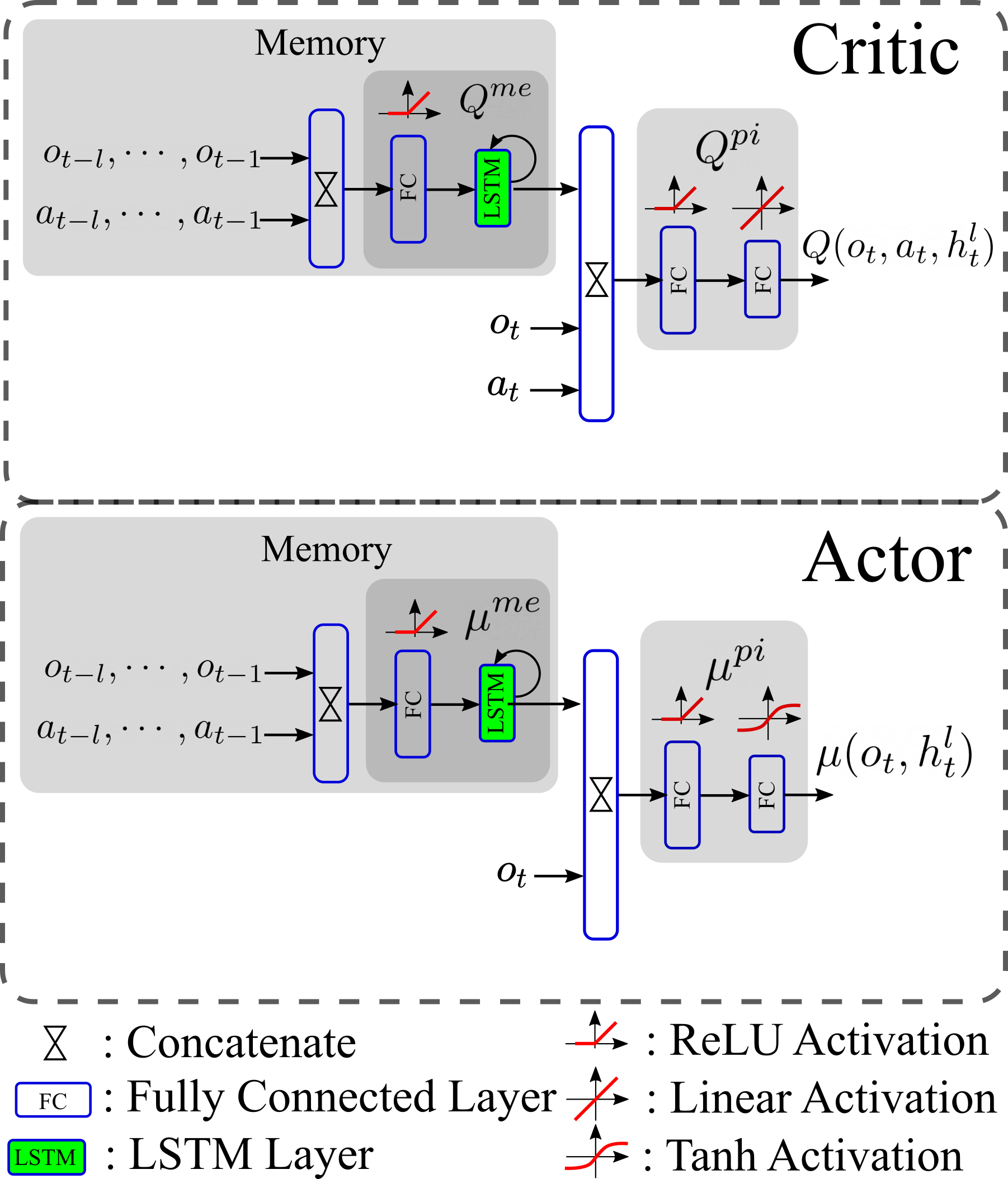}
        \caption{Full$-$CFE}
        \label{fig:Diagram_of_Full_CFE}
    \end{subfigure}
    \hfill
    \begin{subfigure}[t]{.49\linewidth}
        \centering
        \includegraphics[width=\linewidth]{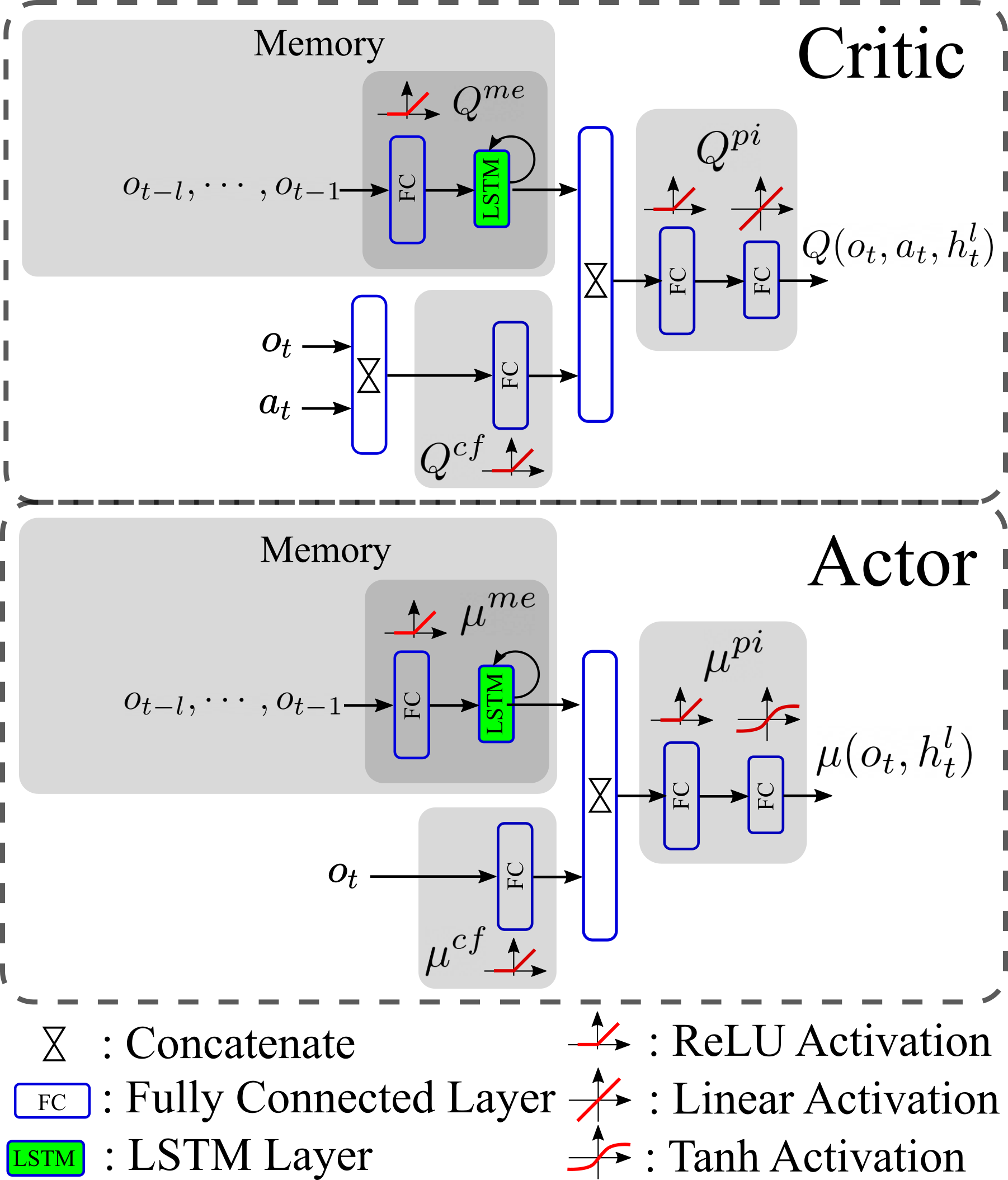}
        \caption{Full$-$PA}
        \label{fig:Diagram_of_Full_PA}
    \end{subfigure}
    \caption{Diagram of Full$-$CFE and Full$-$PA, where for the Full$-$CFE the extracted memory is directly concatenated with the current observation for the actor and with the current observation and action for the critic; and for the Full$-$PA past actions are exclued from the history.}
    \label{fig:Diagram_of_Full_CFE_and_Full_PA}
\end{figure}

\begin{figure*}
    \centering
    \includegraphics[width=.7\linewidth]{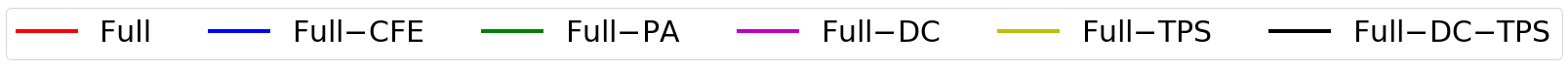}\vspace{3px}
    \subfloat{\includegraphics[width=.99\linewidth]{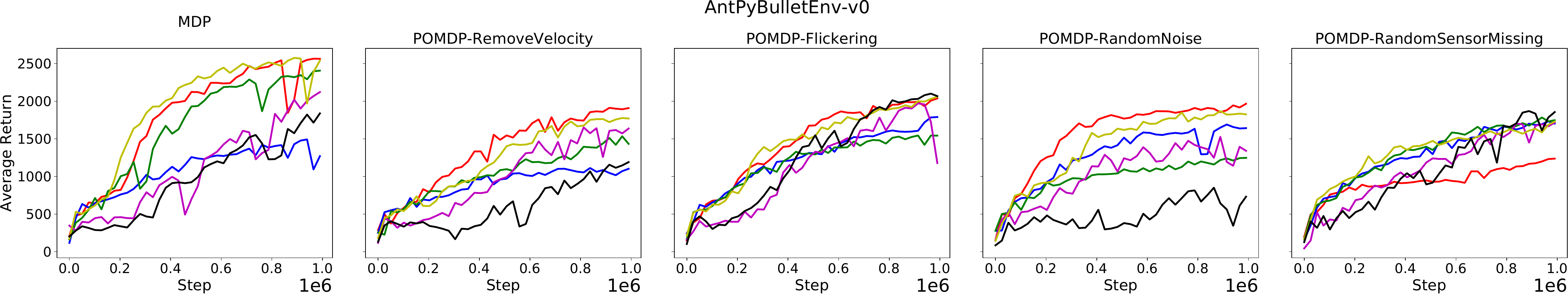}}\\\vspace{5px}
    \subfloat{\includegraphics[width=.99\linewidth]{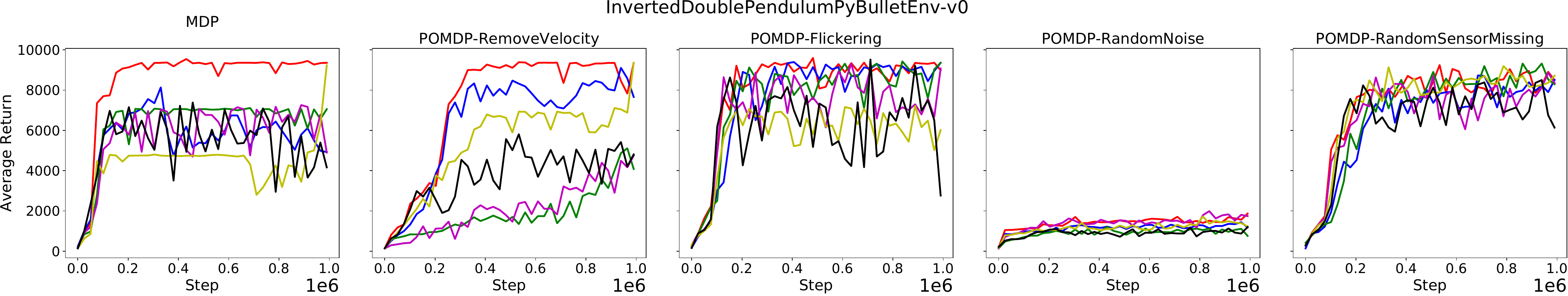}}
    \caption{Learning curves of ablation study, where to ease the comparison only average values over 10 evaluation episodes based on 4 different random seeds are plotted. In the legend, Full, Full$-$CFE, Full$-$PA, Full$-$DC, Full$-$TPS, and Full$-$DC$-$TPS correspond to LSTM-TD3 with full components, removing current feature extraction, excluding past action, not using double critics, not using target policy smoothing, and simultaneously not using double critics and target policy smoothing.}
    \label{fig:learning_curves_for_ablation_performance_comparison}
\end{figure*}

\begin{table}[h]
  \centering
  \caption{Comparing DDPG and LSTM-DDPG in terms of Maximum Average Return, where $\pm$ indicates a single standard deviation. The bolded value of LSTM-DDPG indicates the performance of LSTM-DDPT on a specific version of a task is better than the performance of DDPG on MDP-version of the task.}
  \label{tab:performance_comparison_DDPG_and_LSTM-DDPG}
  \begin{tabular}{crcc}
        \toprule
        \multicolumn{2}{c}{\textbf{Task}}  & \multicolumn{2}{c}{\textbf{Algorithms}}     \\
        \midrule
        Name  &   Version   & \textbf{DDPG}     & \textbf{LSTM-DDPG}  \\
        \midrule
        \multirow{5}{*}{\rotatebox[origin=c]{90}{HalfChePB}}        & MDP       & $487.6\pm6.1$    & \boldmath$517.4\pm102.0$  \\   
                                              & POMDP-RV  & $508.4\pm23.9$   & \boldmath$552.0\pm1.4$  \\  
                                              & POMDP-FLK & $84.8\pm20.4$    & \boldmath$690.8\pm0.0$  \\ 
                                              & POMDP-RN  & $268.7\pm70.2$   & \boldmath$731.1\pm330.9$  \\  
                                              & POMDP-RSM & $283.7\pm27.0$   & \boldmath$606.3\pm63.0$  \\ \hline
        \multirow{5}{*}{\rotatebox[origin=c]{90}{AntPB}}                & MDP       & $1210.8\pm226.1$ & \boldmath$1855.8\pm494.2$ \\ 
                                              & POMDP-RV  & $683.5\pm101.4$  & $1068.6\pm363.0$ \\
                                              & POMDP-FLK & $449.0\pm93.3$   & \boldmath$2145.1\pm107.2$ \\
                                              & POMDP-RN  & $449.6\pm18.5$   & $879.3\pm446.9$  \\
                                              & POMDP-RSM & $465.2\pm51.0$   & \boldmath$1831.7\pm33.9$   \\\hline
        \bottomrule
        \multicolumn{4}{l}{\makecell[l]{$p_{flk}=0.2$, $\sigma_{rn=0.1}$, $p_{rsm}=0.1$.}}
  \end{tabular}
\end{table}

\section{Ablation Study}

To further understand the effect of each component of the proposed LSTM-TD3, in this section we perform an ablation study. Specifically, we examine the effects of the following components: (1) double critics (DC), (2) target policy smoothing (TPS), (3) current feature extraction (CFE), and (4) including past actions (PA) in the history. Fig. \ref{fig:learning_curves_for_ablation_performance_comparison} shows the learning curves of ablated versions of LSTM-TD3, each removing a different component.

\subsection{Effect of Double Critics and Target Policy Smoothing}

As shown in Fig. \ref{fig:learning_curves_for_ablation_performance_comparison}, Full$-$DC shows a significant decrease in performance compared to Full on all MDPs and most POMDPs, whereas TPS seems less important to the best performance of Full. When simultaneously removing DC and TPS, the performance significantly decreases. Note that without DC and the TPS, the LSTM-TD3 is in fact reduced to LSTM-DDPG, similar to RDPG proposed in \cite{heess2015memory}. To ease the comparison, we summarized the results of DDPG and LSTM-DDPG, i.e. Full$-$DC$-$TPS, in Table \ref{tab:performance_comparison_DDPG_and_LSTM-DDPG}. When compared on the same version of a task, LSTM-DDPG always outperforms DDPG, which can be observed by comparing the results in each row. Remarkably, LSTM-DDPG even achieves significantly better performance on POMDPs than DDPG on MDP.

\subsection{Effect of Current Feature Extraction}

In this paper, we intentionally separate the memory extraction and the current feature extraction, then combine them together (Fig. \ref{fig:Recurrent_Actor-Critic_Framework}), in order to differentiate the current and the past and to reduce the interference from useless information in the memory. Alternatively, we can directly combine the current observation for the action (or the current observation and action for the critic) with the extracted memory, i.e. removing the CFE (Full$-$CFE) (Fig. \ref{fig:Diagram_of_Full_CFE}). As shown in Fig. \ref{fig:learning_curves_for_ablation_performance_comparison}, Full$-$CFE performs much worse, compared to Full, especially for MDP-version tasks. Recall that one scenario for devising the LSTM-TD3 is the situation where engineers are not sure if the design of the observation space is appropriate to capture the state of the agent, if the designed observation space properly captures the state of the agent and there is no CFE, poor performance will be achieved. Therefore, CFE is important for such scenarios.

\subsection{Including Past Action Sequence in Memory}
\label{subsec:ablation_add_past_action}

Fig. \ref{fig:Diagram_of_Full_PA} illustrates Full$-$PA, where past actions are excluded from the history. As shown in Fig. \ref{fig:learning_curves_for_ablation_performance_comparison}, removing PA causes a decrease in performance, where a remarkable decrease can be observed on the POMDP-RV version of InvertedDoublePendulumPyBulletEnv-v0 (the 2nd panel in the last row in Fig. \ref{fig:learning_curves_for_ablation_performance_comparison}), which is contrary to the observation in Section \ref{subsec:performance_comparison} that TD3-OW-AddPastAct performs significantly worse than TD3-OW by adding past actions in the history. This means LSTM-TD3 is more robust than OW-TD3. This is desirable especially when designers have no prior about whether observation of past actions is needed to infer the current state of an agent for an unknown task.

\section{Discussion and Conclusion}
In this paper, we proposed a memory-based DRL algorithm called LSTM-TD3 by combining a recurrent actor-critic framework with TD3. The proposed LSTM-TD3 was compared to standard DRL algorithms on both the MDP- and POMDP-versions of continuous control tasks. Our results show that LSTM-TD3 not only achieves significantly better performance on POMDPs than the baselines, but also retains the state-of-art performance on MDP. Our ablation study shows that all components are essential to the success of the LSTM-TD3 where DC and TPS help in stabilizing learning, CFE is especially important to retain the good performance in MDP, and PA is beneficial for tasks where past actions provide information about the current state of the agent.

The proposed approach is particularly useful when engineers do not have enough knowledge about the environment model and the appropriate design of the observation space to capture the underlying state. Memory can be useful in such a scenario to help  infer the underlying state. However, the interpretation of the extracted memory is a challenge. If there is a way to properly interpret the extracted memory, such information, e.g. if the current task is a POMDP or a MDP, can be exploited to improve the observation space design and advance the understanding of the task. Unfortunately, without adding specific constraint terms in the cost functions Eq. \ref{eq:optimize_critic} and \ref{eq:optimize_actor} to facilitate the interpretation, there is no way to properly interpret the extracted memory. Future research should give attention to this direction.

In this paper, for each run of LSTM-TD3 we treat the history length $l$ as a hyper-parameter and fixed it for each run. While LSTM-TD3 with a history length $l=5$ achieves good performance on the devised POMDPs, this may not be achieved for other tasks where the underlying state depends on less recent memory. However, a long history length increases computation resources and time, during both the training and the inferring, i.e. decision making, phases. In the future, an approach for dynamic adaptation of history length $l$ that achieves the best performance while minimising training and decision making time should be investigated. In addition, more sophisticated POMDP tasks relying on long past history should be examined. SAC with LSTM is also worth to investigate in the future. With the insight of the importance of CFE of LSTM-TD3, TD3-OW with a separate CFE should also be studied and compared to LSTM-TD3.





\section*{ACKNOWLEDGMENT}

This research was enabled in part by support provided by Compute Canada (www.computecanada.ca). This work is supported by a SSHRC Partnership Grant in collaboration with Philip Beesley Studio, Inc.

\bibliographystyle{./IEEEtran} 
\bibliography{./IEEEabrv,./root}

\end{document}


\maketitle
\thispagestyle{plain}
\pagestyle{plain}

\begin{abstract}
This manuscript is the supplementary material for the paper "Memory-based Deep Reinforcement Learning for POMDP" by Lingheng Meng, Rob Gorbet and Dana Kuli\'c. It provides additional implementation details, and complementary results. 
\end{abstract}

\section{Algorithms Implementation}
\label{app:Algorithms_Implementation}
The implementation of the algorithms is based on OpenAI Spinningup\footnote{https://spinningup.openai.com}. The code used for this work can be found in \url{https://github.com/LinghengMeng/LSTM-TD3}. Table \ref{tab:Hyperparameters_for_Algorithms} details the hyperparameters used in this work, where $-$ indicates the parameter does not apply to the corresponding algorithm. For the actor and critic neural network structure of LSTM-TD3, the first row corresponds to the structure of the memory component, the second row corresponds to the structure of the current feature extraction, and the third row corresponds to the structure of perception integration after combining the extracted memory and the extracted current feature.

\begin{table}
    \centering
    \caption{Hyperparameters for Algorithms}
    \label{tab:Hyperparameters_for_Algorithms}
    \begin{tabular}{l|c|c|c|c}
        \toprule\hline
        \multirow{2}{*}{\textbf{Hyperparameter}} & \multicolumn{4}{c}{\textbf{Algorithms}}\\\cline{2-5}
            &  \textbf{DDPG} & \textbf{TD3} & \textbf{SAC} & \textbf{LSTM-TD3} \\\hline
        discount factor: $\gamma$  &  \multicolumn{4}{c}{0.99} \\\hline
        batch size: $N_{batch}$    &  \multicolumn{4}{c}{100}  \\\hline
        \makecell[l]{replay buffer size: \\$\left | D \right |$} &  \multicolumn{4}{c}{$10^{6}$}  \\\hline
        \makecell[l]{random start step: \\$N_{start\_step}$}    &  \multicolumn{4}{c}{$10000$}  \\\hline
        \makecell[l]{update after \\$N_{update\_after}$}  &  \multicolumn{4}{c}{$1000$}  \\\hline
        \makecell[l]{target NN update \\rate $\tau$}               &  \multicolumn{4}{c}{$0.005$}  \\\hline
        optimizer & \multicolumn{4}{c}{Adam \cite{kingma2014adam}} \\\hline
        \makecell[l]{actor learning rate \\$lr_{actor}$}          &  \multicolumn{4}{c}{$10^{-3}$}  \\\hline
        \makecell[l]{critic learning rate \\$lr_{critic}$}         &  \multicolumn{4}{c}{$10^{-3}$}  \\\hline
        \multirow{3}{*}{actor NN structure:}  & \multicolumn{3}{c|}{\multirow{3}{*}{[256, 256]}}  & $[128]+[128]$\\
        &\multicolumn{1}{c}{} &\multicolumn{1}{c}{} & & $[128]$\\
        &\multicolumn{1}{c}{} &\multicolumn{1}{c}{} & & $[128, 128]$\\\hline
        \multirow{3}{*}{critic NN structure:} & \multicolumn{3}{c|}{\multirow{3}{*}{[256, 256]}} & $[128]+[128]$\\
        &\multicolumn{1}{c}{} &\multicolumn{1}{c}{} & & $[128]$\\
        &\multicolumn{1}{c}{} &\multicolumn{1}{c}{} & & $[128, 128]$\\\hline
        \makecell[l]{actor exploration \\noise $\sigma_{act}$}       &  \multicolumn{2}{c|}{$0.1$} & - & $0.1$ \\\hline
        \makecell[l]{target actor noise \\$\sigma_{targ\_act}$} & - & $0.2$ & - & $0.2$ \\\hline
        \makecell[l]{target actor noise clip \\boundary $c_{targ\_act}$}      & - & $0.5$ & - & $0.5$ \\\hline
        policy update delay  & - & 2 & - & 2 \\\hline
        \makecell[l]{entropy regulation \\coefficient $\alpha$}            & - & - & 0.2 & - \\\hline
        history length $l$                                 & - & - & -& \{0, 1, 3, 5\}   \\\hline
        \bottomrule
    \end{tabular}
    
\end{table}

\begin{figure*}
    \centering
    \includegraphics[width=.5\linewidth]{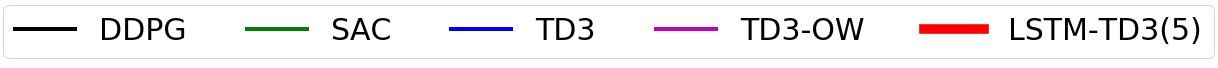}
    \subfloat[POMDP-RN]{\includegraphics[width=.49\linewidth]{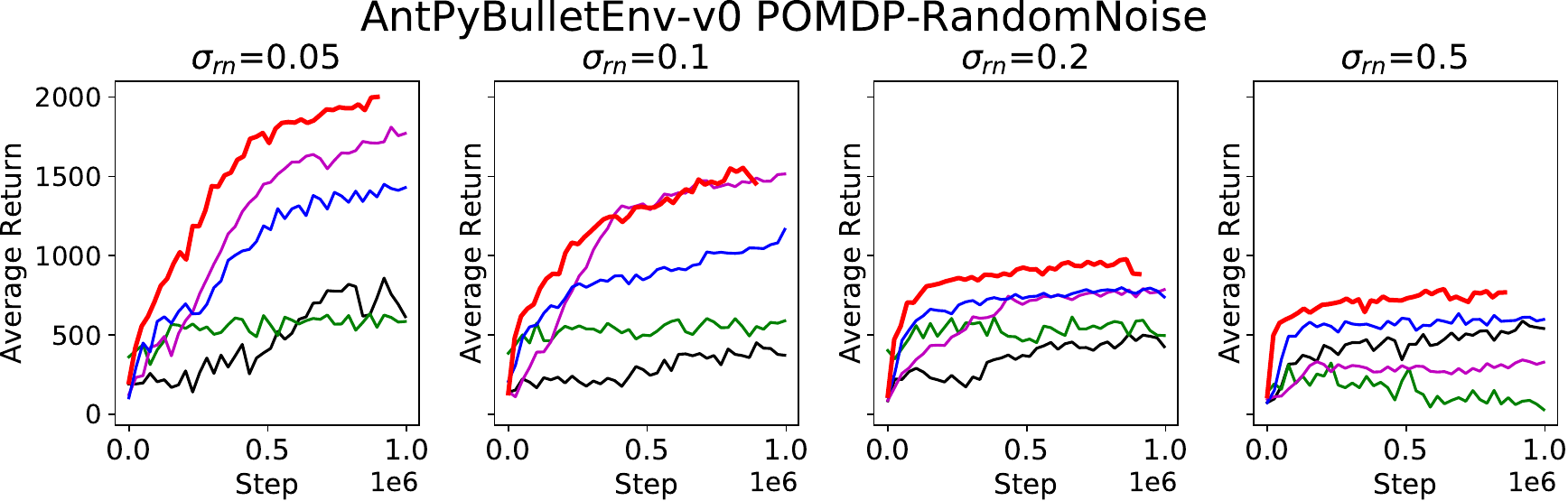}}
    \subfloat[POMDP-RSM]{\includegraphics[width=.49\linewidth]{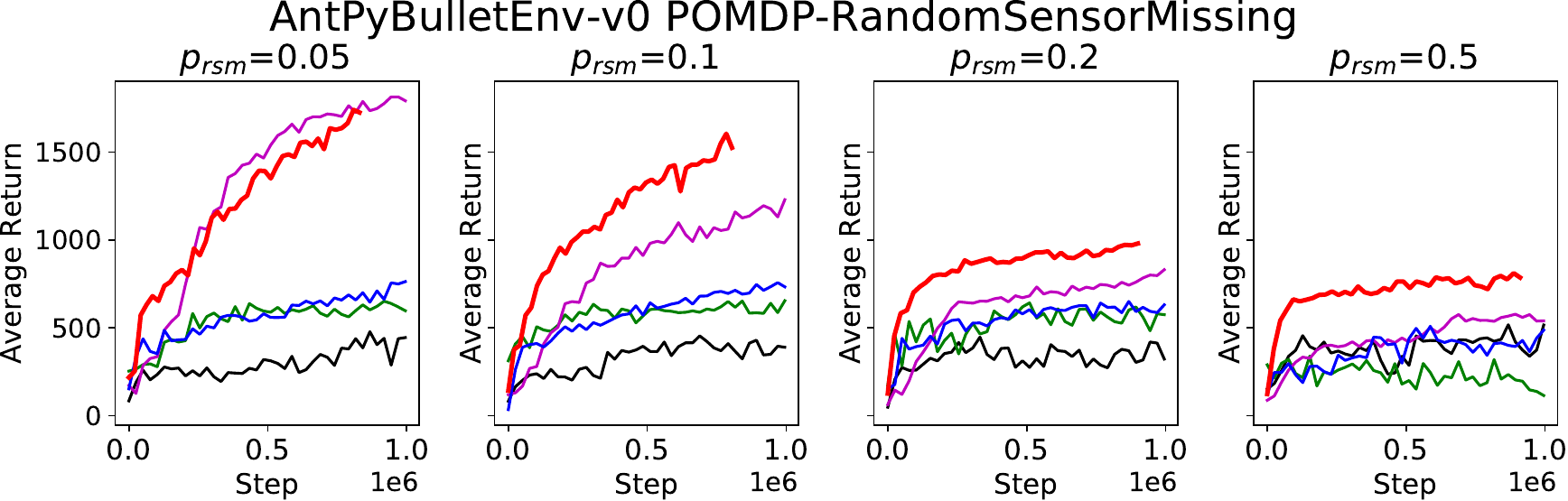}}
    \caption{Performance Comparison on POMDP-version of AntPyBulletEnv-v0 with different observabilities.}
    \label{fig:robustness_to_partially_observability}
\end{figure*}
\begin{figure}
    \centering
    \includegraphics[width=.99\linewidth]{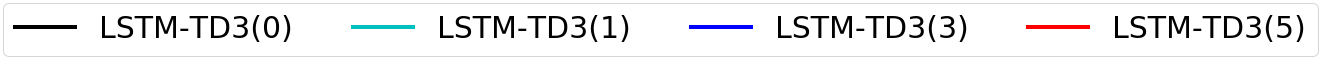}
    \includegraphics[width=.99\linewidth]{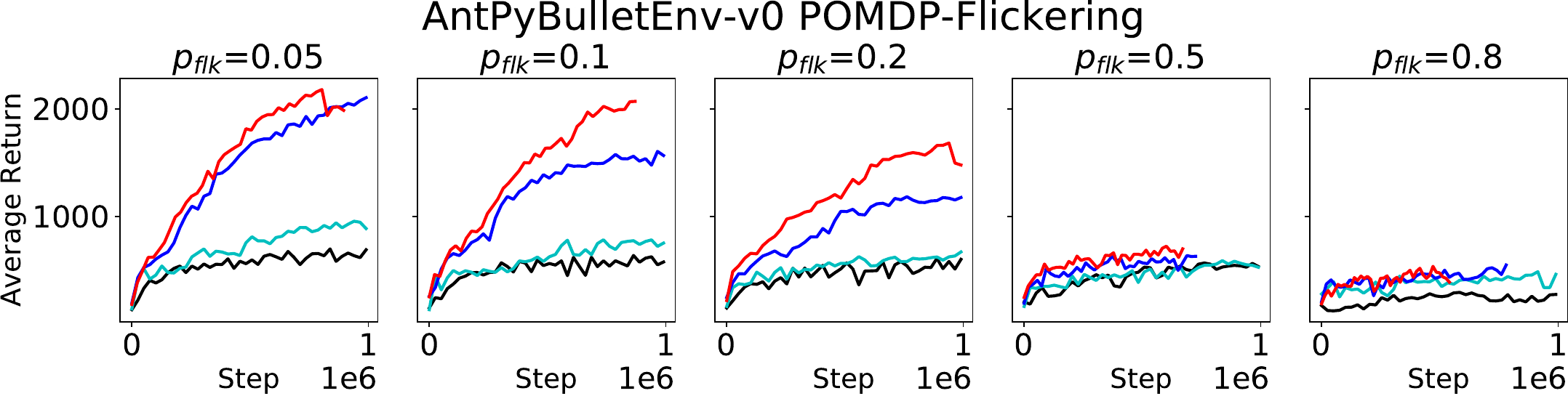}
    \caption{Relationship Between Partial-Observability and History Length.}
    \label{fig:Relationship_Between_Partially_Observability_and_History_Length}
\end{figure}

\begin{figure*}
    \centering
    \includegraphics[width=.99\linewidth]{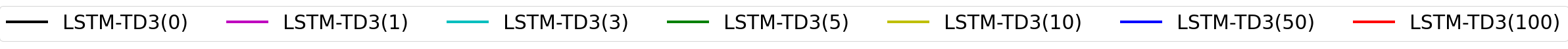}\vspace{2px}
    \includegraphics[width=.24\linewidth]{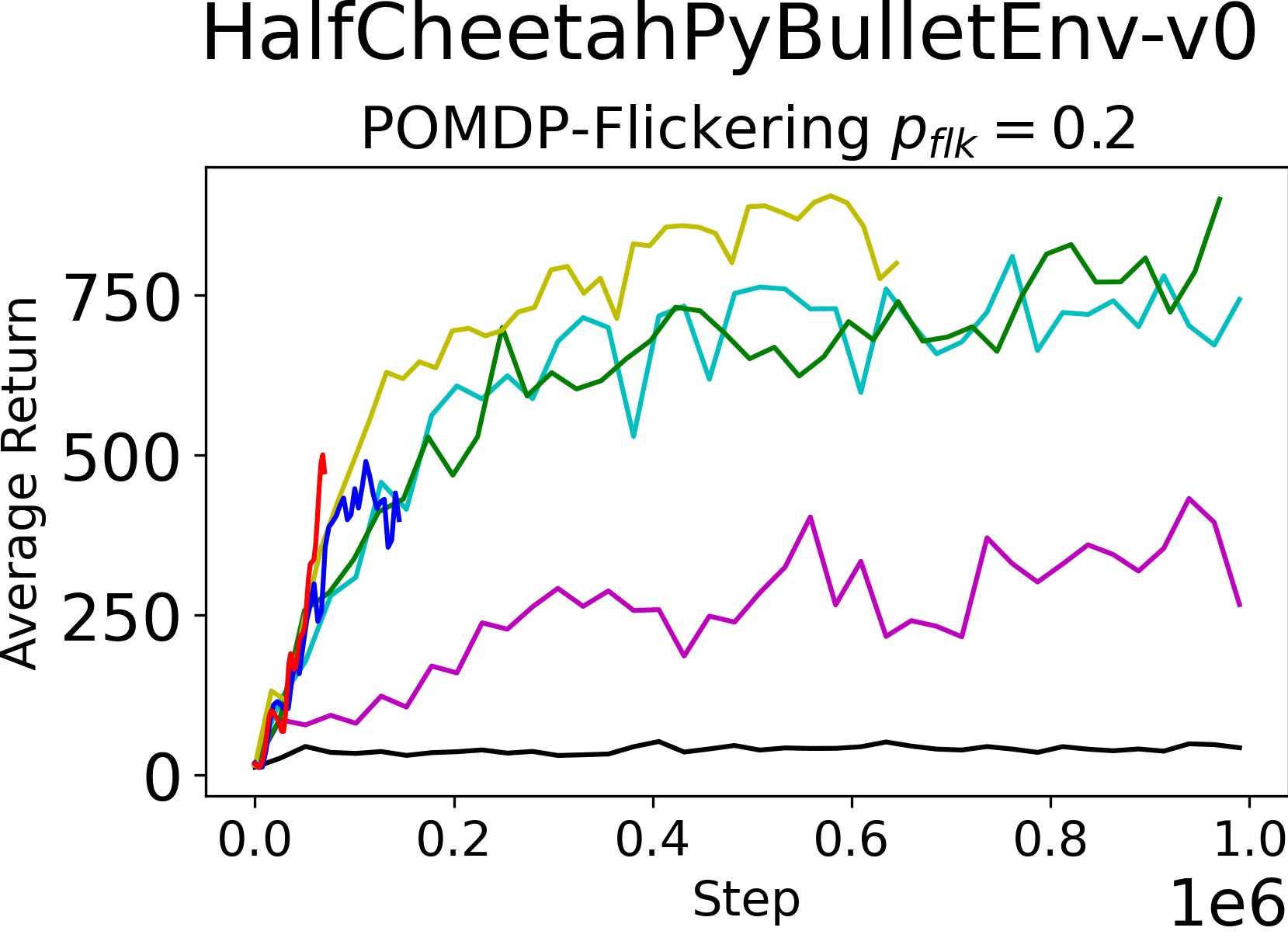}
    \includegraphics[width=.24\linewidth]{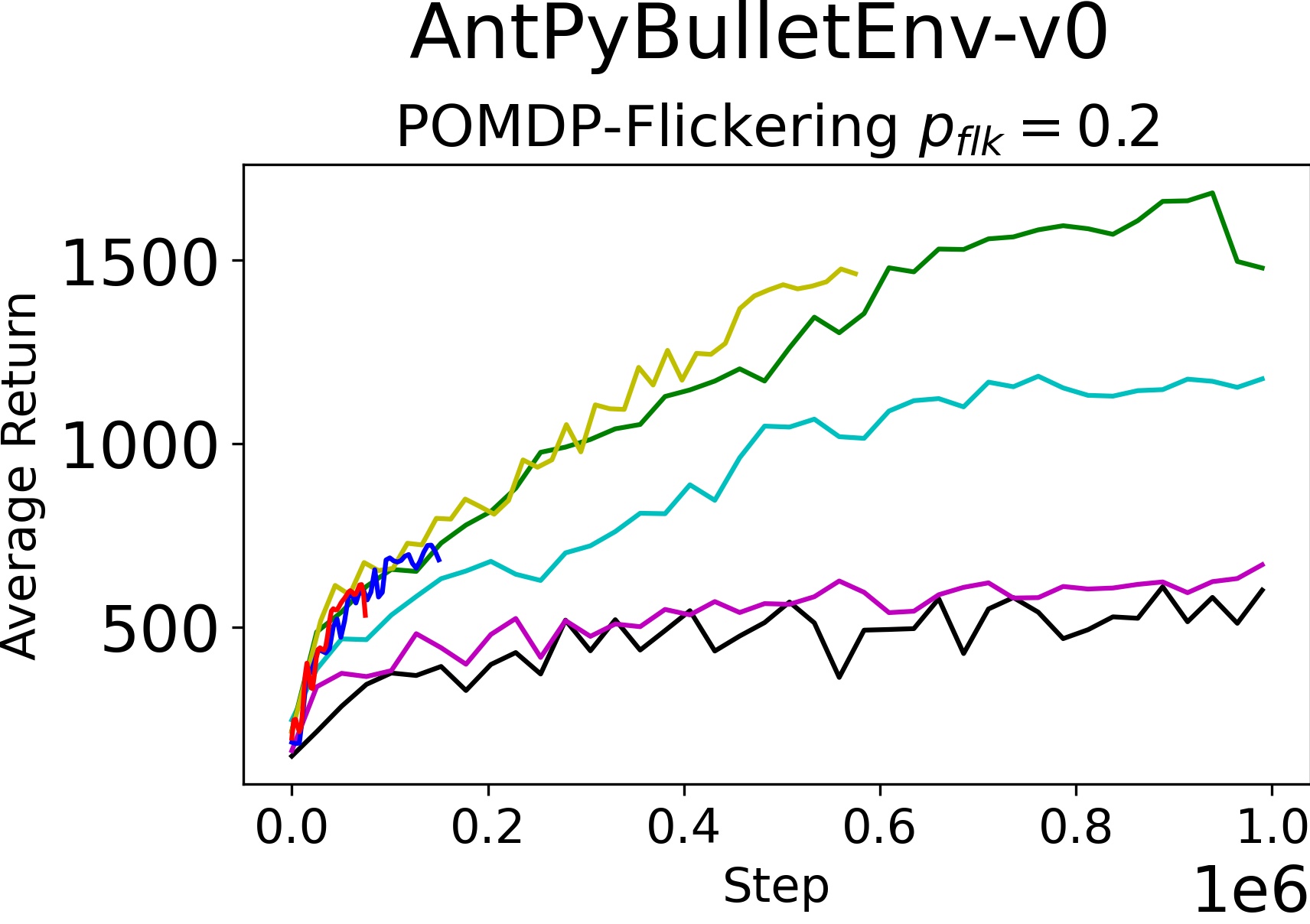}
    \includegraphics[width=.24\linewidth]{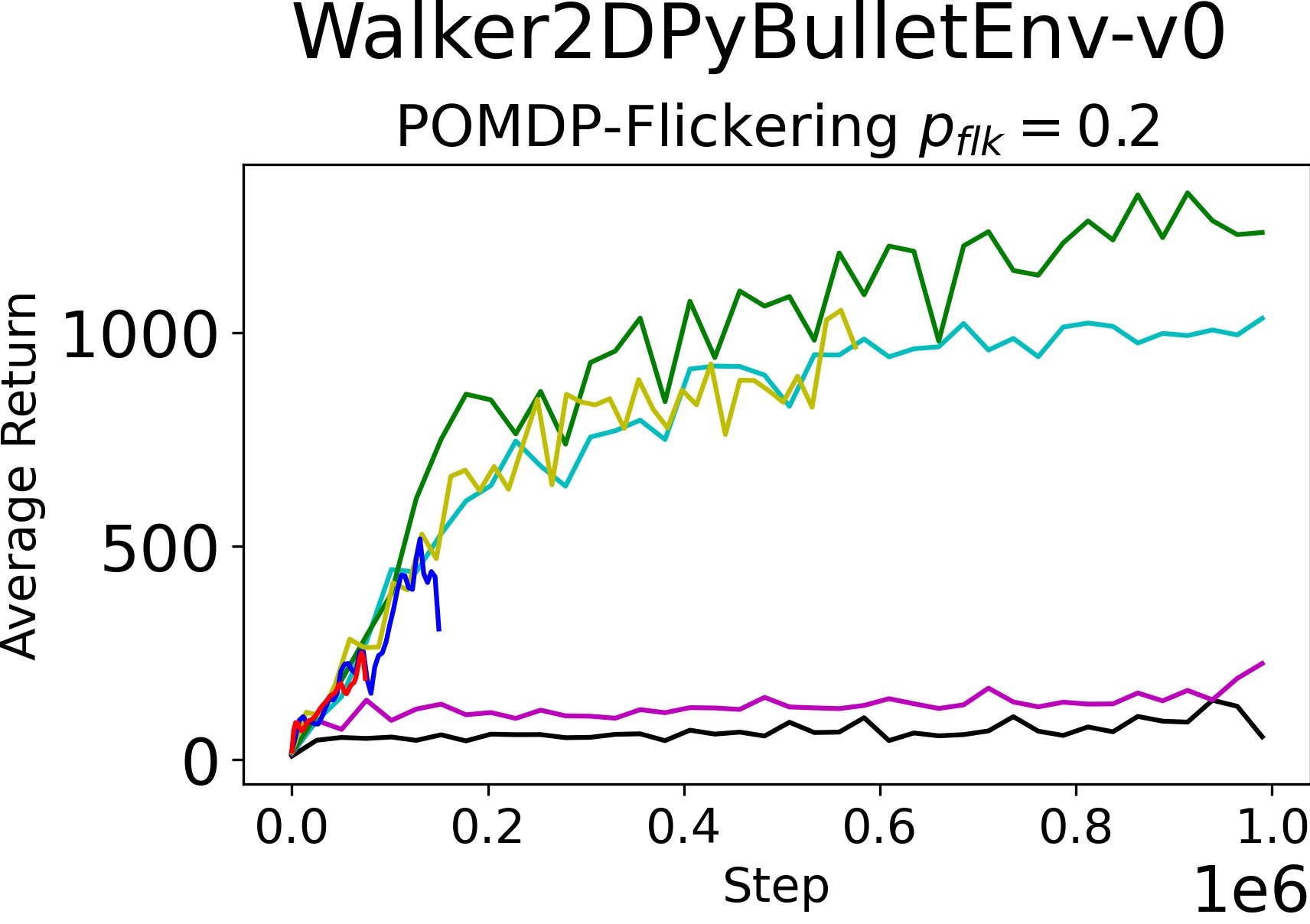}
    \includegraphics[width=.24\linewidth]{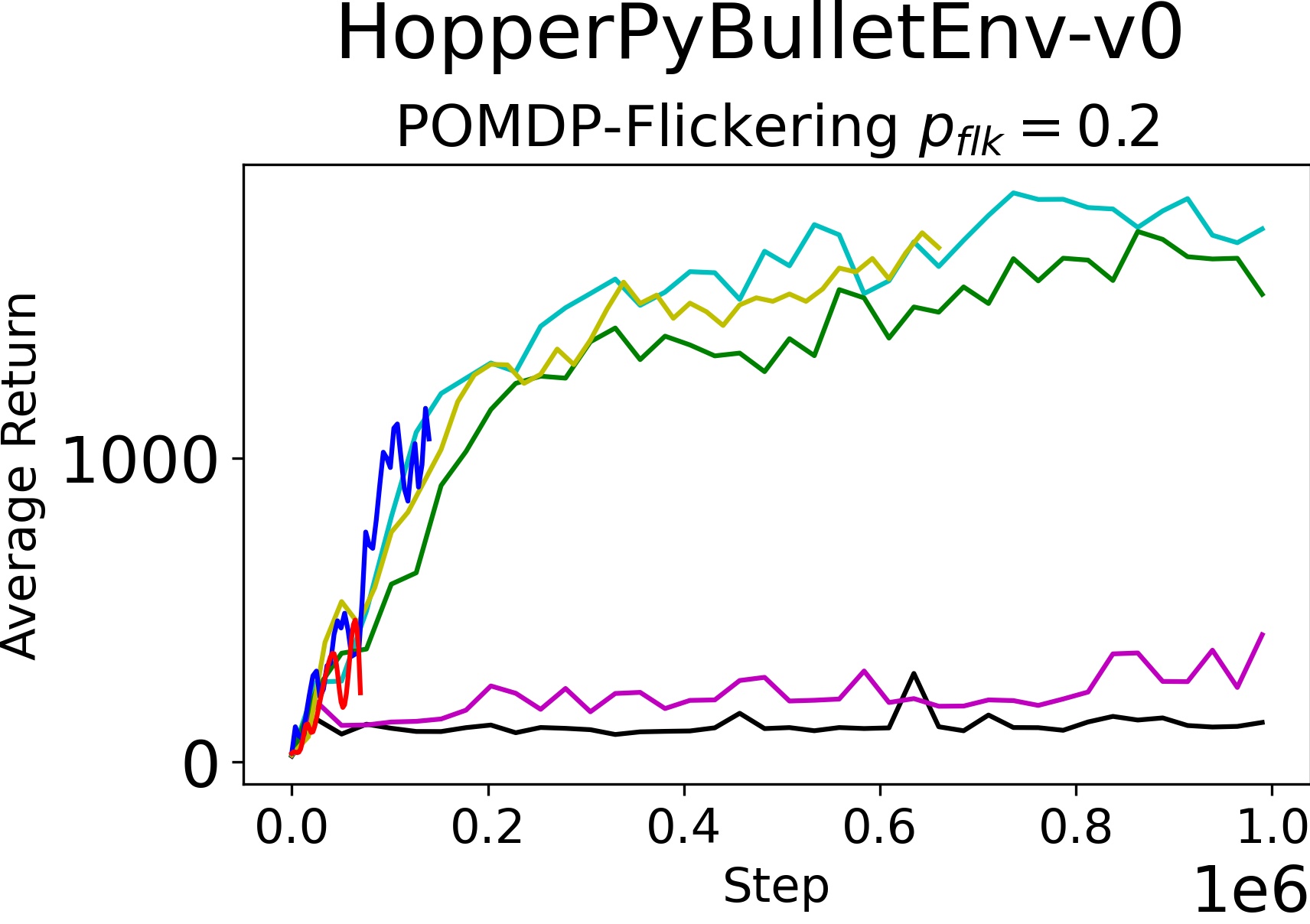}
    \caption{Performance of LSTM-TD3 with Different History Lengths, where LSTM-TD3 with history length 10, 50 and 100 are not fully run up to 1 million steps due to the extra computation cost caused by long history.}
    \label{fig:performance_for_different_History_Lengths}
\end{figure*}

\section{Supplementary Results}
\subsection{Performance Comparison}
\label{subsec:performance_comparison}
\begin{figure*}[t]
    \centering
    \includegraphics[width=.65\linewidth]{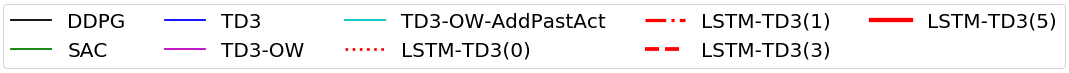}\vspace{5px}
    \subfloat{\includegraphics[width=.99\linewidth]{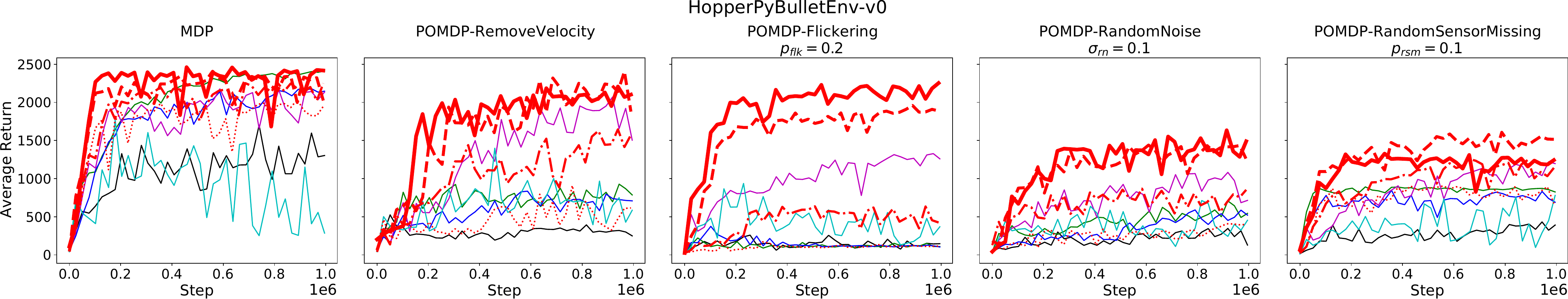}}\vspace{5px}
    \subfloat{\includegraphics[width=.99\linewidth]{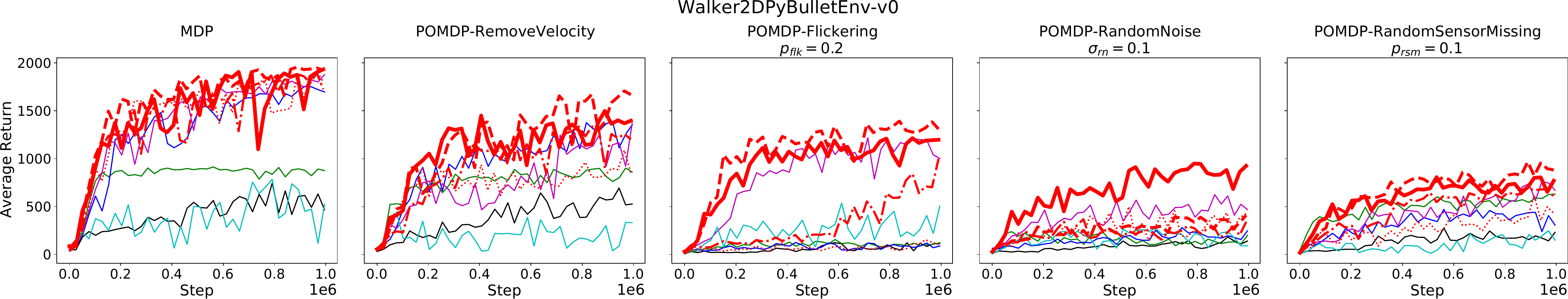}}\vspace{5px}
    \subfloat{\includegraphics[width=.99\linewidth]{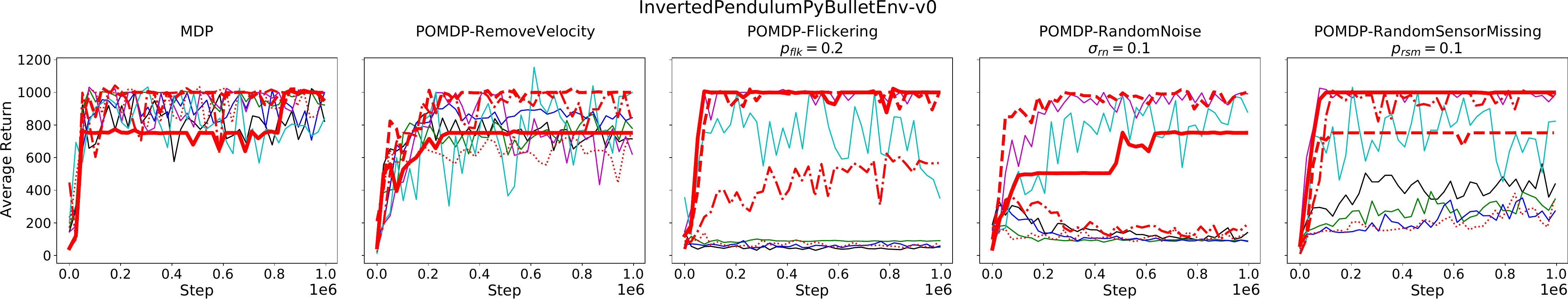}}\vspace{5px}
    \caption{Learning curves for PyBulletGym tasks, where to ease the comparison only average values are plotted. In the legend, the value in the bracket of LSTM-TD3 indicate the length of the history, e.g. LSTM-TD3(5) uses the history length 5.}
    \label{fig:learning_curves_for_performance_comparison}
\end{figure*}

Table \ref{tab:performance_comparison} summarizes the maximum average return of each algorithm on different tasks, where the POMDP-FLK, POMDP-RN, and POMDP-RSM are examined with $p_{flk}=0.2$, $\sigma_{rn}=0.1$, and $p_{rsm}=0.1$. From Table \ref{tab:performance_comparison}, we can see that LSTM-TD3 either outperforms or achieves comparative performance to other baselines on MDP, and significantly outperforms other baselines on POMDP-versions of each task. These results provide evidence of the promising advantages of memory based DRL on both MDP and POMDP. On one hand, LSTM-TD3 can be used out-of-box without caring too much of the design of the observation as memory component can partially compensate the lost information. On the other hand, by comparing the performance difference of adding and removing the memory component, LSTM-TD3 provides a way to detect if the current design of the observation space is improvable or not in terms of capturing the underlying state.

\begin{table*}[h]
  \centering
  \caption{Maximum Average Return over 10 evaluation episodes based on 4 different random seeds. Maximum value of evaluated algorithms for each task is bolded, and $\pm$ indicates a single standard deviation. LSTM-TD3 (5) corresponds to the LSTM-TD3 with the history length $l=5$.}
  \label{tab:performance_comparison}
  \begin{tabular}{rrccccc}
        \toprule
        \multicolumn{2}{c}{\textbf{Task}}  & \multicolumn{5}{c}{\textbf{Algorithms}}     \\
        \midrule
        Name  &   Version   & \textbf{DDPG}     & \textbf{TD3} & \textbf{SAC} & \textbf{TD3-OW} & \textbf{LSTM-TD3 (5)} \\
        \midrule
        \multirow{5}{*}{HalfCheetahPB}  & MDP & $487.6\pm6.1$ & \boldmath{$1311.9\pm49.7$} & $663.5\pm30.8$ & $1265.5\pm8.9$ & {$1223.0\pm582.3$} \\   
                                              & POMDP-RV & $508.4\pm23.9$ & $1151.7\pm74.9$ & $631.5\pm57.1$ & \boldmath$1161.3\pm17.2$ & $918.4\pm44.0$ \\  
                                              & POMDP-FLK & $84.8\pm20.4$ & $82.4\pm45.7$ & $117.0\pm42.2$ & \boldmath$1559.61\pm559.9$ & {$848.1\pm60.2$} \\ 
                                              & POMDP-RN & $268.7\pm70.2$ & $501.9\pm47.5$ & $328.4\pm62.1$ & $703.9\pm21.7$ & \boldmath{$771.6\pm18.2$} \\  
                                              & POMDP-RSM & $283.7\pm27.0$ & $538.9\pm32.2$ & $587.4\pm44.3$ & $606.9\pm13.4$ & \boldmath{$954.0\pm362.9$} \\ \hline
        \multirow{5}{*}{AntPB}                & MDP & $1210.8\pm226.1$ & $2433.5\pm288.5$ & $980.8\pm96.3$ & $2289.5\pm154.8$ & \boldmath{$2574.9\pm79.0$} \\ 
                                              & POMDP-RV & $683.5\pm101.4$ & $1765.6\pm2.2$ & $800.4\pm4.8$ & $1265.3\pm65.2$ & \boldmath$1932.7\pm199.6$ \\
                                              & POMDP-FLK & $449.0\pm93.3$ & $654.4\pm1.6$ & $529.7\pm23.7$ & $1390.5\pm736.6$ & \boldmath$2036.7\pm73.5$ \\
                                              & POMDP-RN & $449.6\pm18.5$ & $1165.8\pm59.0$ & $620.8\pm10.0$ & $1520.1\pm8.4$ & \boldmath$1966.1\pm171.4$ \\
                                              & POMDP-RSM & $465.2\pm51.0$ & $763.7\pm103.3$ & $659.1\pm3.1$ & $1230.4\pm124.0$ & \boldmath$1324.9\pm313.6$  \\\hline
        \multirow{5}{*}{Walker2DPB}           & MDP & $835.0\pm102.2$ & $1783.1\pm111.6$ & $930.1\pm53.2$ & $1941.3\pm128.5$ & \boldmath$1970.5\pm38.5$ \\
                                              & POMDP-RV & $716.6\pm224.5$ & $1477.9\pm164.3$ & $921.9\pm20.8$ & $1220.9\pm53.8$ & \boldmath$1479.9\pm283.2$ \\
                                              & POMDP-FLK & $142.4\pm29.6$ & $181.9\pm98.7$ & $217.2\pm90.6$ & $1238.6\pm385.4$ & \boldmath$1264.9\pm338.5$ \\
                                              & POMDP-RN & $197.2\pm96.2$ & $295.8\pm44.7$ & $278.9\pm44.5$ & $648.3\pm129.5$ & \boldmath$984.7\pm267.7$ \\
                                              & POMDP-RSM & $283.6\pm31.0$ & $519.4\pm17.5$ & $630.5\pm20.8$ & $633.2\pm23.2$ & \boldmath$841.2\pm91.6$ \\ \hline
        \multirow{5}{*}{HopperPB}             & MDP & $1699.6\pm80.2$ & $2201.3\pm180.4$ & $2424.5\pm85.4$ & $2210.1\pm286.4$ & \boldmath$2465.0\pm158.9$ \\
                                              & POMDP-RV & $520.6\pm105.3$ & $926.0\pm219.6$ & $1145.8\pm162.1$ & $2212.1\pm5.5$ & \boldmath$2233.6\pm176.6$ \\
                                              & POMDP-FLK & $259.5\pm63.9$ & $401.1\pm39.8$ & $243.2\pm161.3$ & $1353.0\pm467.8$ & \boldmath$2264.6\pm72.3$ \\
                                              & POMDP-RN & $400.8\pm62.8$ & $644.2\pm46.5$ & $782.0\pm65.2$ & $962.0\pm10.5$ & \boldmath$1635.8\pm180.7$ \\
                                              & POMDP-RSM & $596.1\pm57.1$ & $873.2\pm7.9$ & $892.3\pm2.5$ & $1193.7\pm193.3$ & \boldmath$1349.1\pm405.7$ \\\hline
        \multirow{5}{*}{InvPendulumPB}        & MDP & \boldmath$1000.0\pm0.0$ & \boldmath$1000.0\pm0.0$ & \boldmath$1000.0\pm0.0$ & \boldmath$1000.0\pm0.0$ & \boldmath$1000.0\pm0.0$ \\
                                              & POMDP-RV & $912.5\pm13.5$ & \boldmath$944.1\pm41.6$ & $891.3\pm108.7$ & $876.3\pm123.7$ & $752.6\pm428.5$ \\
                                              & POMDP-FLK & $147.1\pm120.1$ & $158.1\pm115.5$ & $121.1\pm23.0$ & \boldmath$1000.0\pm0.0$ & \boldmath$1000.0\pm0.0$ \\ 
                                              & POMDP-RN & $422.5\pm18.8$ & $342.8\pm46.8$ & $289.8\pm19.3$ & \boldmath$1000.0\pm0.0$ & $752.6\pm428.5$ \\
                                              & POMDP-RSM & $624.9\pm4.0$ & $381.5\pm180.1$ & $445.0\pm101.5$ & \boldmath$1000.0\pm0.0$ & \boldmath$1000.0\pm0.0$ \\\hline
        \multirow{5}{*}{InvDoublePenPB}  & MDP & $4746.3\pm4607.1$ & $7054.3\pm2304.5$ & $9357.3\pm0.3$ & $9358.6\pm0.3$ & \boldmath$9359.77\pm0.1$ \\
                                              & POMDP-RV & $750.6\pm130.4$ & $953.6\pm92.0$ & $2027.6\pm508.6$ & $7998.1\pm653.7$ & \boldmath$9358.9\pm0.3$ \\
                                              & POMDP-FLK & $274.2\pm62.8$ & $382.6\pm105.4$ & $404.9\pm11.8$ & \boldmath$9358.6\pm0.5$ & $9358.4\pm1.1$ \\ 
                                              & POMDP-RN & $506.5\pm159.3$ & $470.4\pm330.3$ & $662.7\pm34.7$ & \boldmath$2005.3\pm13.4$ & $1952.7\pm503.4$ \\
                                              & POMDP-RSM & $881.7\pm364.4$ & $829.5\pm96.1$ & $1084.9\pm103.6$ & \boldmath$9304.4\pm54.6$ & $9156.1\pm348.6$ \\
        \bottomrule
        \multicolumn{7}{l}{For POMDP-FLK, $p_{flk}=0.2$. For POMDP-RN, $\sigma_{rn=0.1}$. For POMDP-RSM, $p_{rsm}=0.1$.}
  \end{tabular}
\end{table*}

\subsection{Robustness to Partial Observability}

Fig. \ref{fig:robustness_to_partially_observability} compares the proposed LSTM-TD3 with the baselines in terms of the robustness to different partial observabilities, where the higher the $\sigma_{rn}$ and the $p_{rsm}$, the lower the observability. In general, LSTM-TD3 has better robustness than TD3-OW which has better robustness than TD3. A small reduction of observability can be handled by LSTM-TD3, but for POMDP with severe reduction of observability LSTM-TD3 performance also degrades.

\subsection{Effect of History Length}

The history length $l$ is a hyperparameter of LSTM-TD3, and it determines the maximum  history length in the observation window. As illustrated in Fig. \ref{fig:learning_curves_for_performance_comparison} and Table \ref{tab:performance_comparison}, LSTM-TD3 with a relatively short history length $l=5$ produces significantly better performance than other baselines without a memory component. Fig. \ref{fig:Relationship_Between_Partially_Observability_and_History_Length} shows the performance of LSTM-TD3 with different history lengths on POMDP-FLK AntPyBulletEnt-v0 with various flickering probabilities $p_{flk}=\{0.05,0.1,0.2,0.5,0.8\}$ where the higher the $p_{flk}$ the lower the obervability. For $p_{flk}=0.05$, LSTM-TD3(3) and LSTM-TD3(5) show similar performance and both significantly outperform LSTM-TD3(0) and LSTM-TD3(1). When $p_{flk}$ increases from 0.05 to 0.1, LSTM-TD3(5) still maintains similar performance, but LSTM-TD3(3) experiences a dramatic decrease. This means the decreased observability can still be compensated with history of length 5, but cannot be compensated with history of length 3. When further reducing the observability to $p_{flk}=0.2$, the performance of LSTM-TD3(5) is also degraded, as shown in the 3rd panel of Fig. \ref{fig:Relationship_Between_Partially_Observability_and_History_Length}. When $p_{flk}$ is increased to 0.5 and 0.8, all examined history lengths fail the task (the last two panels in Fig. \ref{fig:Relationship_Between_Partially_Observability_and_History_Length}).

Fig. \ref{fig:performance_for_different_History_Lengths} illustrates the performance of LSTM-TD3 with different history lengths. From this figure, we can see that when the history length increased, the final performance improves too. However, the long history length causes much extra computation consumption. 

\subsection{Policy Generalization}
The second extension is to evaluate the learned policy with different history length from that used during training. If we use $l_{train}$ and $l_{eval}$ to represent the history length used during training and evaluation respectively, the generalization capability for both $l_{train}>l_{eval}$ and $l_{train}<l_{eval}$ can be useful in different scenarios. On one hand, for $l_{train}>l_{eval}$, if $l_{train}$ and $l_{eval}$ can achieve the same performance, using a shorter history length during evaluation can reduce the inferring time of an action, and this is valuable for tasks where real-time decision making is important and training can be run in parallel and is not time-sensitive. On the other hand, for $l_{train}<l_{eval}$, if $l_{eval}$ can achieve better performance than $l_{train}$ by increasing the history length, which means how to extract useful memory information can be learned with a shorter history length and longer history length during evaluation is only to extract more useful information, then using a shorter history length during training can speed up the training and reduce resource consumption, at the same time without compromising the performance. As for other DRL algorithms, the training is on a mini-batch while the evaluating is only on a single data, so if the $l_{train}$ is large, the training will take more time and computation resources than evaluation. Therefore, it a good choice to achieve the same performance by reducing the cost during training and increasing the cost during evaluation. Normally, training cost is unvalued by researchers, especially some research \cite{adamski2018distributed,kalashnikov2018qt} proposed to use large distributed computer cluster. However, there are cases, where the robot cannot communicate fleetly with the remote computation center and onboard computation resources are limited, that the effort in saving computation is still required. This practical consideration inspires the second evaluation extension.

\begin{figure*}
    \centering
    \includegraphics[width=.59\linewidth]{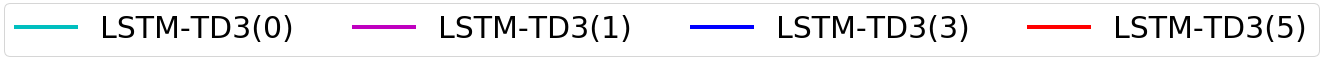}
    \includegraphics[width=.89\linewidth]{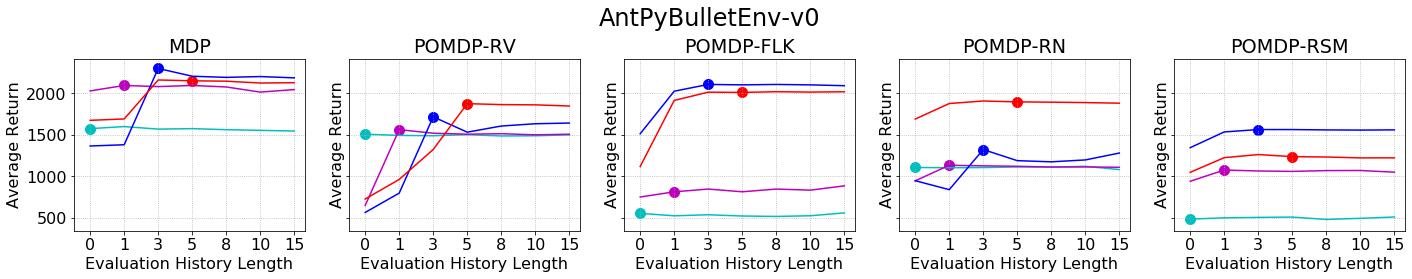}
    \caption{Evaluation with History Length Different From that Used When Training, where in each panel the title indicates the version of the task, the x-axis corresponds to the history length used during evaluation, each line corresponds to the policy trained with a specific history length, and the marker indicates the point where the training and evaluation history length are the same.}
    \label{fig:evaluation_on_different_history_length}
\end{figure*}

\subsubsection{Evaluation with Different History Lengths}
\label{subsubsec:Evaluation_with_Different_History_Lengths}
Fig. \ref{fig:evaluation_on_different_history_length} illustrates the evaluation results of LSTM-TD3 with various history lengths that may be different from the history length used for training. From this figure, it can be seen that when trained with a specific history length $l_{train}$ but evaluated with a different history length $l_{eval} > l_{train}$, the performance remains at the same level. However, when evaluated with a history $l_{eval} < l_{train}$, the performance cannot be guaranteed, as for some cases a shorter evaluation history length will cause dramatic decrease in performance, e.g. LSTM-TD3(3) on POMDP-RN (the blue line in the 4th panel of Fig. \ref{fig:evaluation_on_different_history_length}), while for others a shorter evaluation history length can still achieve similar performance, e.g. LSTM-TD(5) on POMDP-RN and on POMDP-RSM (the red line in the 4th and 5th panels of Fig. \ref{fig:evaluation_on_different_history_length}). Based on this observation, it seems the performance can be generalized to a longer evaluation history length rather than a shorter one. It is worth to note that the longest training history length investigate here is only 5, so more valuable insights may be found with more results with longer training history length.

\begin{figure*}
    \centering
    \includegraphics[width=.59\linewidth]{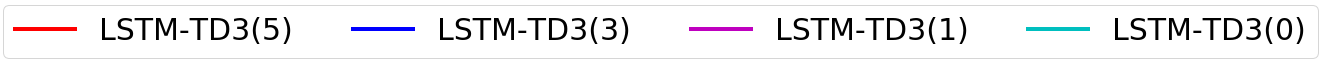}\vspace{5px}
    \subfloat[Average Test Return]{\includegraphics[width=.99\linewidth]{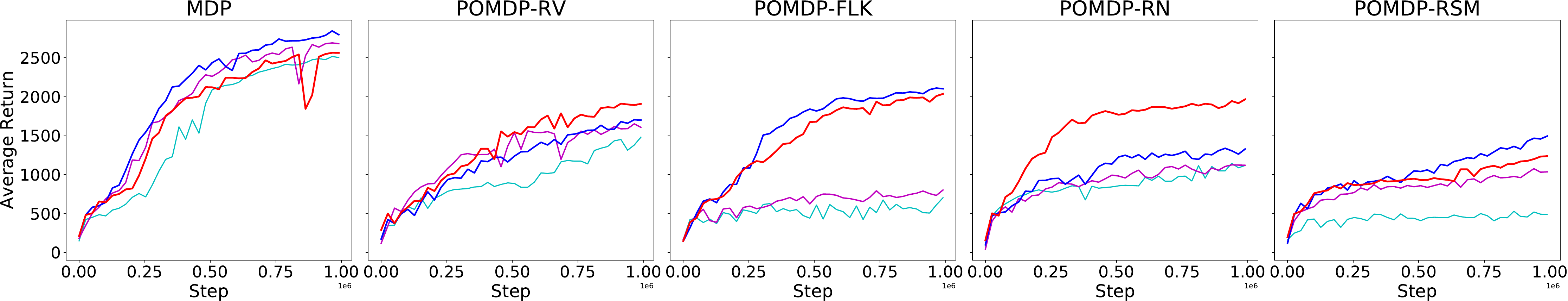}\label{fig:Analysis_AverageTestEpRet_AntPyBulletEnv-v0}}\\\vspace{5px}
    \subfloat[Average Q Value]{\includegraphics[width=.99\linewidth]{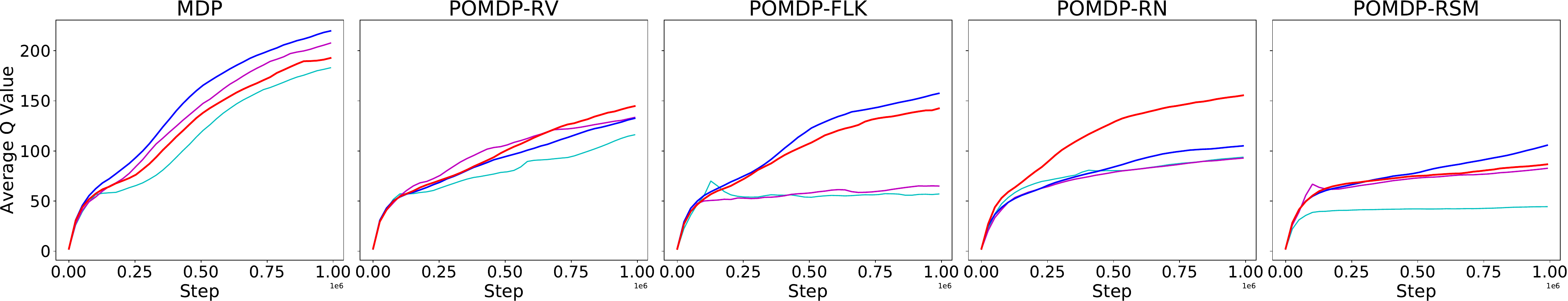} \label{fig:Analysis_AverageQ1Vals_AntPyBulletEnv-v0}}\\\vspace{5px}
    \subfloat[Average Extracted Memory of Actor]{\includegraphics[width=.99\linewidth]{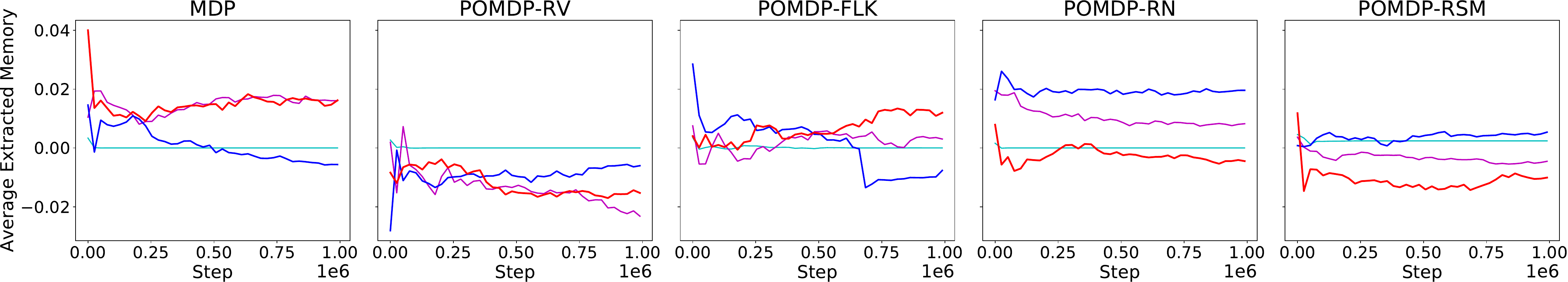}\label{fig:Analysis_AverageActExtractedMemory_AntPyBulletEnv-v0}}\vspace{5px}
    \subfloat[Average Extracted Memory of Critic]{\includegraphics[width=.99\linewidth]{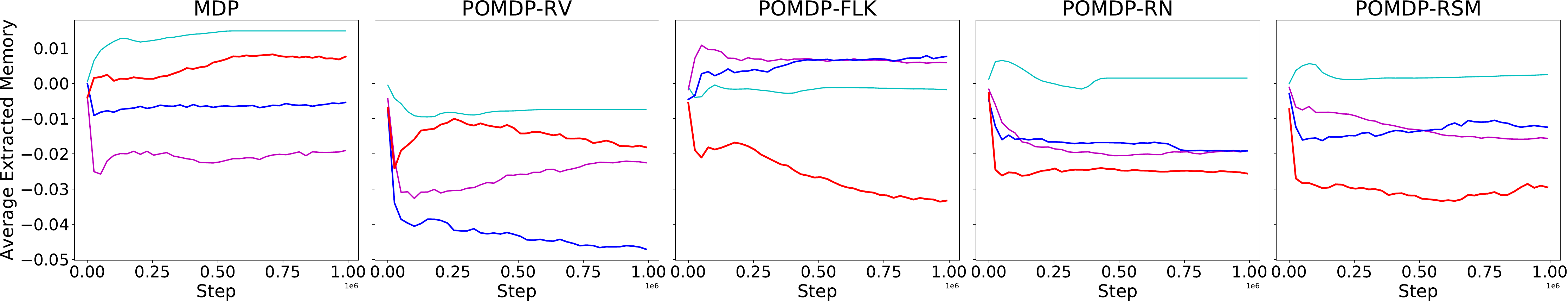}\label{fig:Analysis_AverageQ1ExtractedMemory_AntPyBulletEnv-v0}}
    \caption{Relationship Among the Return, Predicted Q-value, Extracted Memory of Actor-Critic, where (a), (b), (c), and (d) shows the average test return, the average predicted Q-value, the average extracted memory of actor, and the average extracted memory of critic. The average extracted memory of actor and critic is an average over all output neurons of the memory component.}
    \label{fig:Relationship_Among_the_Return_Predicted_Q_value_Extracted_Memory_of_Actor_Critic}
\end{figure*}

The results that when evaluating with history length longer than 0, the performance does not change (the cyan lines in Fig. \ref{fig:evaluation_on_different_history_length}) of LSTM-TD3(0), are very interesting, because in the training phase LSTM-TD3(0) takes zero-valued dummy observation and action as history as defined in Eq. \ref{eq:history_definition} and this dummy history cannot provide any useful information about how to extracting memory that is useful to current task. Therefore, when replacing this dummy history with real history in the replay buffer, the extracted memory can be anything and will disturb the decision making, which makes us to expect that when evaluating the learned policy with history length longer than 0, the performance will be decreased. However, the results is surprisingly remained at the same level as that for LSTM-TD3(0). This reminds us that LSTM-TD3(0) may have learned a policy that intentionally ignores the history. To validate this conjecture, we plotted the average extracted memory of the actor in Fig. \ref{fig:Analysis_AverageActExtractedMemory_AntPyBulletEnv-v0}. As shown in Fig. \ref{fig:Analysis_AverageActExtractedMemory_AntPyBulletEnv-v0}, the average extracted memory of the actor of LSTM-TD3(0) initially starts with a non-zero value, but after a few thousands steps its value remains at a value around $0.0034 \pm 0.0024$, which is very close to 0, whereas for other LSTM-TD3s with longer history length the average extracted memories are relatively far away from 0 and have relatively large standard deviation. And this observation is consistent for both MDP and POMDPS. Even though we cannot claim the 0 in the extracted memory can be interpreted as neglect of past history, but at least we can say the history is uniformly mapped to a roughly fixed value rather than a random value for each history. In this way, even replacing the zero-value dummy history with a real history of experiences, the performance will not bad than that for a dummy history.

\begin{figure*}
    \centering
    \includegraphics[width=.9\linewidth]{images/ablation_study/ablation_legend.png}\vspace{3px}
    \subfloat{\includegraphics[width=.99\linewidth]{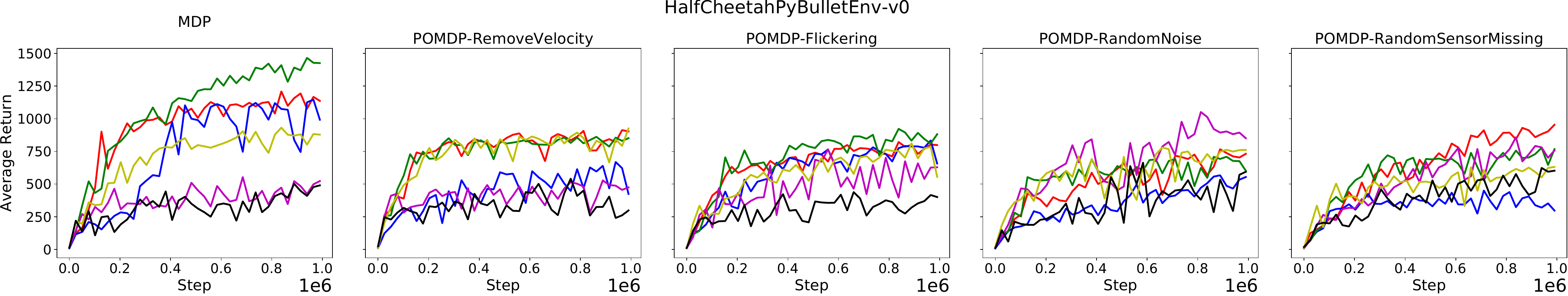}}\\\vspace{5px}
    \subfloat{\includegraphics[width=.99\linewidth]{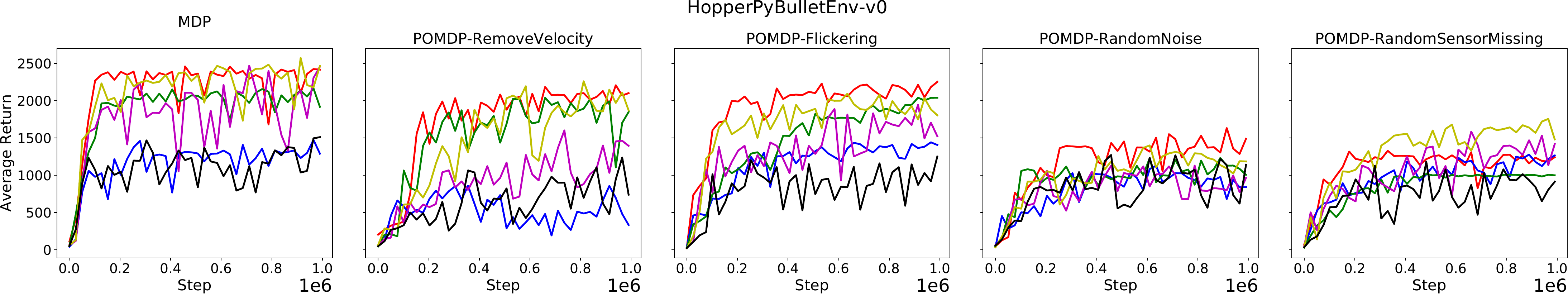}}\vspace{5px}
    \subfloat{\includegraphics[width=.99\linewidth]{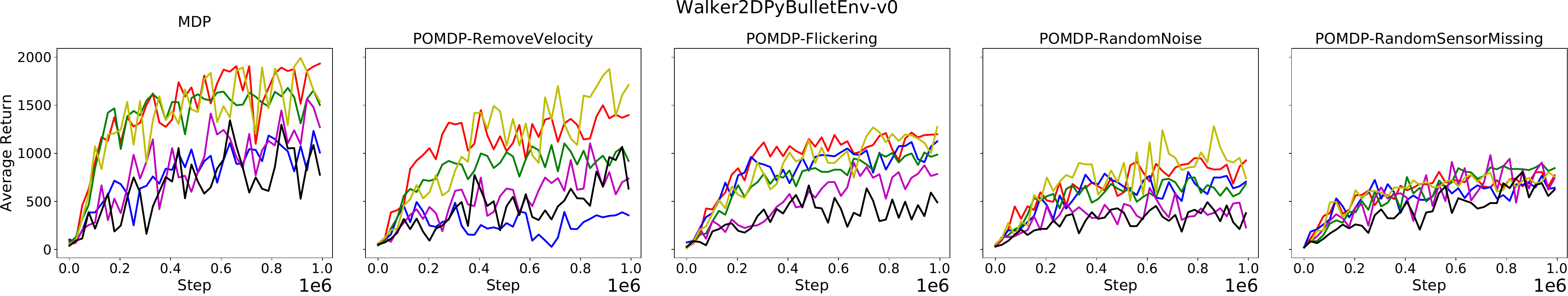}}\vspace{5px}
    \caption{Learning curves of ablation study, where to ease the comparison only average values are plotted. In the legend, Full, Full$-$CFE, Full$-$PA, Full$-$DC, Full$-$TPS, and Full$-$DC$-$TPS correspond to LSTM-TD3 with full components, removing current feature extraction, excluding past action, not using double critics, not using target policy smoothing, and simultaneously not using double critics and target policy smoothing.}
    \label{fig:learning_curves_for_ablation_performance_comparison_complement}
\end{figure*}

\begin{table*}[h]
  \centering
  \caption{Maximum Average Return for Ablation Study. Maximum value of evaluated algorithms for each task is bolded, and $\pm$ indicates a single standard deviation.}
  \label{tab:ablation_performance_comparison}
  \begin{tabular}{crcccccc}
        \toprule
        \multicolumn{2}{c}{\multirow{2}{*}{\textbf{Task}}}  & \multicolumn{6}{c}{\textbf{Algorithms}}     \\ 
        \cmidrule{3-8}
                     &   & \textbf{LSTM-TD3}    & \textbf{Full$-$CFE} & \textbf{Full$-$PA}  & \textbf{Full$-$DC} & \textbf{Full$-$TPS} & \textbf{Full$-$DC$-$TPS}  \\
        \midrule
        \multirow{5}{*}{\rotatebox[origin=c]{90}{HalfCheePB}}  & MDP       & {$1223.0\pm582.3$}         & $1182.2\pm522.0$ & \boldmath$1469.6\pm637.5$ & $569.1\pm78.5$ & $920.6\pm48.8$ & $517.4\pm102.0$ \\   
                                        & POMDP-RV  & $918.4\pm44.0$    & $680.6\pm129.9$ & $871.0\pm43.6$ & $589.5\pm73.8$  & \boldmath$945.6\pm187.5$ & $552.0\pm1.4$ \\  
                                        & POMDP-FLK & {$848.1\pm60.2$}           & $844.7\pm211.3$ & \boldmath$946.9\pm242.1$& $836.3\pm346.7$ & $809.9\pm42.7$ & $690.8\pm0.0$ \\ 
                                        & POMDP-RN  & {$771.6\pm18.2$}           & $641.5\pm40.2$  & $742.9\pm79.2$ & \boldmath$1222.2\pm313.4$ & $808.3\pm468.1$ & $731.1\pm330.9$ \\  
                                        & POMDP-RSM & $954.0\pm362.9$ & $457.0\pm99.1$  & $784.0\pm39.7$ & \boldmath$958.1\pm229.1$ & $693.9\pm79.0$ & $606.3\pm63.0$ \\ \hline
        \multirow{5}{*}{\rotatebox[origin=c]{90}{AntPB}}          & MDP       & {$2574.9\pm79.0$} & $1510.8\pm421.4$ & $2421.0\pm305.4$ & $2121.1\pm334.6$ & \boldmath$2658.6\pm1537.1$ & $1855.8\pm494.2$ \\ 
                                        & POMDP-RV  & \boldmath$1932.7\pm199.6$  & $1133.2\pm421.6$ & $1535.0\pm358.1$ & $1711.9\pm331.5$ & $1814.3\pm90.3$ & $1068.6\pm363.0$\\
                                        & POMDP-FLK & $2036.7\pm73.5$   & $1817.2\pm323.0$ & $1578.1\pm466.2$ & $2023.9\pm348.0$ & $2083.8\pm159.9$ & \boldmath$2145.1\pm107.2$ \\
                                        & POMDP-RN  & \boldmath$1966.1\pm171.4$  & $1705.7\pm105.8$ & $1287.4\pm366.0$ & $1588.0\pm548.0$ & $1885.8\pm92.7$ & $879.3\pm446.9$ \\
                                        & POMDP-RSM & $1324.9\pm313.6$  & $1737.4\pm310.3$ & $1817.6\pm66.4$  & $1728.8\pm494.0$ & $1730.0\pm474.4$ & \boldmath$1831.7\pm33.9$  \\\hline
        \multirow{5}{*}{\rotatebox[origin=c]{90}{Walker2DPB}}     & MDP       & $1970.5\pm38.5$  & $1230.9\pm329.2$ & $1719.6\pm431.1$ & $1526.3\pm96.5$  & \boldmath$2011.7\pm61.6$ & $1381.9\pm801.0$ \\
                                        & POMDP-RV  & $1479.9\pm283.2$ & $520.2\pm85.4$   & $1100.9\pm243.6$ & $1214.2\pm313.7$ & \boldmath$1871.7\pm138.3$ & $1349.3\pm101.1$ \\
                                        & POMDP-FLK & $1264.9\pm338.5$ & $1132.6\pm256.2$ & $1027.3\pm214.5$ & $1017.1\pm122.6$ & \boldmath$1320.2\pm275.9$ & $731.5\pm160.0$ \\
                                        & POMDP-RN  & $984.7\pm267.7$  & $844.9\pm282.2$  & $803.5\pm22.9$   & $662.7\pm155.2$  & \boldmath$1240.7\pm228.2$ & $576.6\pm403.8$ \\
                                        & POMDP-RSM & $841.2\pm91.6$   & $762.9\pm53.0$   & $896.1\pm46.8$   & \boldmath$1070.8\pm226.4$ & $820.8\pm45.4$ & $1010.2\pm324.9$ \\ \hline
        \multirow{5}{*}{\rotatebox[origin=c]{90}{HopperPB}}       & MDP       & $2465.0\pm158.9$ & $1517.0\pm673.8$ & $2178.2\pm200.7$ & $2504.7\pm180.5$ & \boldmath$2613.1\pm81.5$ & $2015.0\pm477.6$ \\
                                        & POMDP-RV  & \boldmath$2233.6\pm176.6$ & $1187.0\pm749.1$ & $2188.9\pm240.2$ & $1775.5\pm471.6$ & $2303.9\pm156.6$ & $1852.9\pm327.4$ \\
                                        & POMDP-FLK & \boldmath$2264.6\pm72.3$  & $1459.0\pm513.2$ & $2055.9\pm48.1$ & $2087.9\pm194.0$ & $2117.9\pm103.1$ & $2083.3\pm168.7$ \\
                                        & POMDP-RN  & \boldmath$1635.8\pm180.7$ & $1143.9\pm298.7$ & $1279.7\pm444.7$ & $1337.7\pm262.1$ & $1450.7\pm357.3$ & $1524.0\pm516.0$ \\
                                        & POMDP-RSM & $1349.1\pm405.7$ & $1343.1\pm411.3$ & $1008.6\pm3.4$ & $1634.3\pm338.6$ & $1769.5\pm210.6$ & \boldmath$1938.2\pm0.0$ \\\hline
        \multirow{5}{*}{\rotatebox[origin=c]{90}{InvPenPB}}       & MDP       &  \boldmath$1000.0\pm0.0$  & $874.2\pm217.9$ & \boldmath$1000.0\pm0.0$ & \boldmath$1000.0\pm0.0$ & \boldmath$1000.0\pm0.0$ & \boldmath$1000.0\pm0.0$  \\
                                        & POMDP-RV  & $752.6\pm428.5$  & $804.4\pm246.2$ & $752.6\pm428.6$ & \boldmath$1000.0\pm0.0$ & \boldmath$1000.0\pm0.0$ & \boldmath$1000.0\pm0.0$  \\
                                        & POMDP-FLK & \boldmath$1000.0\pm0.0$   & $1000.0\pm0.0$  & \boldmath$1000.0\pm0.0$ & \boldmath$1000.0\pm0.0$ & \boldmath$1000.0\pm577.4$ & \boldmath$1000.0\pm0.0$  \\ 
                                        & POMDP-RN  & $752.6\pm428.5$  & $752.7\pm428.3$ & \boldmath$1000.0\pm0.0$ & \boldmath$1000.0\pm0.0$ & \boldmath$1000.0\pm0.0$ & $986.5\pm19.0$ \\
                                        & POMDP-RSM & \boldmath$1000.0\pm0.0$   & \boldmath$1000.0\pm0.0$ & \boldmath$1000.0\pm0.0$ & \boldmath$1000.0\pm0.0$ & \boldmath$1000.0\pm0.0$ & \boldmath$1000.0\pm0.0$  \\\hline
        \multirow{5}{*}{\rotatebox[origin=c]{90}{InvDouPenPB}} & MDP       & \boldmath$9359.73\pm0.1$     & $8677.9\pm749.7$ & $9359.0\pm5403.4$ & $9343.9\pm5394.7$ & $9358.8\pm9358.8$ & $9352.5\pm5399.7$ \\
                                        & POMDP-RV  & \boldmath$9358.9\pm0.3$   & $ 9358.0\pm1.0$ & $5568.0\pm4941.9$ & $4748.1\pm6616.0$ & $9357.7\pm5402.7$ & $8974.0\pm377.3$ \\
                                        & POMDP-FLK & $9358.4\pm1.1$    & $9358.0\pm1.1$ & $9357.4\pm1.9$ & $9356.2\pm1.2$ & $9357.1\pm5402.3$ & \boldmath$9359.4\pm0.0$  \\ 
                                        & POMDP-RN  & $1952.7\pm503.4$  & $1534.0\pm462.1$ & $1308.5\pm1220.5$ & $2453.9\pm1576.2$ & $2360.5\pm1403.6$ & \boldmath$2544.9\pm3599.0$ \\
                                        & POMDP-RSM & $9156.1\pm348.6$  & $ 9139.5\pm377.1$ & \boldmath$9358.7\pm0.5$ & $9156.9\pm333.3$ & $9357.7\pm0.8$ & $9355.0\pm3.8$ \\
        \bottomrule
        \multicolumn{8}{l}{For POMDP-FLK, $p_{flk}=0.2$. For POMDP-RN, $\sigma_{rn=0.1}$. For POMDP-RSM, $p_{rsm}=0.1$.}\\
        \multicolumn{8}{l}{CFE: Current Feature Extraction; PA: Past Action; DC: Double Critics; TPS: Target Policy Smooth}
  \end{tabular}
\end{table*}

\subsection{A Glance of The Relationship Among the Return, the Predicted Q-value, and the Extracted Memory of the Actor-Critic}
\label{subsec:a_glance_of_the_relationship}

The proposed LSTM-TD3 has been experimentally proved to be useful according to the results presented in Section \ref{subsec:performance_comparison}, but the understanding of the LSTM-TD3, especially the interpretation of the extracted memory, is still a big challenge. In Fig. \ref{fig:Relationship_Among_the_Return_Predicted_Q_value_Extracted_Memory_of_Actor_Critic}, we presented the average test return, the average predicted Q-value, the average extracted memory of actor and critic to have a glance of the relationship among them. 

By comparing Fig. \ref{fig:Analysis_AverageTestEpRet_AntPyBulletEnv-v0} and \ref{fig:Analysis_AverageQ1Vals_AntPyBulletEnv-v0}, we found that the average test return and the average predicted Q-value match approximately perfect, which means there is no overestimation problem as studied in \cite{thrun1993issues,fujimoto2018addressing,meng2020effect} and is desirable. 

Our special interest is in the relationship between the test return and the extracted memory of the actor-critic. It is worth to note that both the actor and the critic have a memory component and there is no sharing of the memory component, which means they may learn different coding of the memory that is helpful for learning a Q-value function and a policy, respectively. As shown in Fig. \ref{fig:Analysis_AverageActExtractedMemory_AntPyBulletEnv-v0} and \ref{fig:Analysis_AverageQ1ExtractedMemory_AntPyBulletEnv-v0}, no matter if we comparing these plots horizontally or vertically, there is no consistent trend can be found. 
Horizontally, for both the actor and the critic, the average extracted memories for different versions, i.e. the MDP and the various POMDPs, of the task are showing different trends, where an exception is the extracted memory of the actor of LSTM-TD3(0) as discussed in the previous section \ref{subsubsec:Evaluation_with_Different_History_Lengths}. Vertically, for different versions of the task, the extracted memories of the actor and the critic show different trends too. Especially, the average extracted memory of the critic of the LSTM-TD3(0) is not similar with that of the actor whose value is consistently close to 0. Contrarily, the average extracted memory of the critic of LSTM-TD3(0) is even further from 0 than that of LSTM-TD3(5). This may indicate the different roles of the memory component playing in the actor and the critic. Again, the interpretation of the extracted memory is not so straight-forward, and special constraints may be forced to improve the interpretability of the memory component, which is out of the scope of this paper and will be left for the future study.

\subsection{Supplementary results for the Ablation Study}

Fig. \ref{fig:learning_curves_for_ablation_performance_comparison_complement} shows the learning curves of  various  ablated  LSTM-TD3  by  removing  different  components, namely (1) using double critics (DC), (2) using target policy smoothing (TPS), (3) having current feature extraction (CFE) component, and (4) including past actions (PA) in the history. Table \ref{tab:ablation_performance_comparison} reports the maximum average return of the investigated ablated algorithms.

\bibliographystyle{./IEEEtran} 
\bibliography{./IEEEabrv,./root}


\clearpage
